\documentclass[pmlr]{jmlr}

\usepackage{times}

\usepackage[utf8]{inputenc}
\usepackage[T1]{fontenc}
\usepackage{amsfonts}
\usepackage{bm}
\usepackage{color}

\usepackage{mathrsfs}
\usepackage{relsize}
\usepackage{appendix}
\usepackage{algorithmic}
\usepackage{algorithm}

\usepackage[final]{showlabels} % final to mute it
\showlabels{cite}

\usepackage[disable]{todonotes} % hide notes

\renewcommand{\leq}{\leqslant}
\renewcommand{\geq}{\geqslant}

\newcommand{\lint}{[\![}
\newcommand{\rint}{]\!]}
\newcommand{\intint}[2]{\lint #1 , #2 \rint}
\newcommand{\cond}{\,|\,}
\newcommand{\midd}{\,\Vert\,}

\newcommand{\kll}[2]{\Delta ( {#1} \midd {#2} )} % Kullback-Leibler divergence

\newcommand{\pp}{\, : \; }

\newcommand{\N}{\mathbf N}

\newcommand{\R}{\mathbf R}

\newcommand{\X}{\mathscr X}
\newcommand{\Y}{\mathscr Y}
\newcommand{\M}{\mathscr{M}}

\newcommand{\pr}{\mathscr{P}}

\renewcommand{\ln}{\log}
\newcommand{\ie}{\textit{i.e.} }
\newcommand{\eg}{e.g. }

\newcommand{\Pweight}{\Pi}

\newcommand{\Pw}{\Pweight}

\newcommand{\fseq}{\mathscr{S}^{(f)}}
\newcommand{\aseq}{\mathscr{S}^{(a)}}
\newcommand{\sseq}{\mathscr{S}^{(s)}}

\makeatletter
\g@addto@macro\normalsize{%
  \setlength\abovedisplayskip{5pt}
  \setlength\belowdisplayskip{5pt}
  \setlength\abovedisplayshortskip{5pt}
  \setlength\belowdisplayshortskip{5pt}
}
\makeatother

\newtheorem{assumption}{\bf Assumption}%[section]% numérotation
\newtheorem{rem}{\bf Remark}

\newcommand{\BlackB}{\rule{1.5ex}{1.5ex}}
\newcommand*{\myqed}{\hfill\BlackB%\\[2mm]
  \medskip
}
\newenvironment{myproof}%
{%
 \par\noindent{\bfseries\upshape Proof\ }%
}%
{\myqed}

\newcommand{\ghedge}{{\textcolor{red!60!black}{GrowingHedge}}}
\newcommand{\mhedge}{{\textcolor{red!60!black}{MarkovHedge}}}
\newcommand{\fmhedge}{\textcolor{red!60!black}{FreshMarkovHedge}}
\newcommand{\gmhedge}{\textcolor{red!60!black}{GrowingMarkovHedge}}
\newcommand{\smhedge}{\textcolor{red!60!black}{SleepingMarkovHedge}}
\newcommand{\gsmhedge}{\textcolor{red!60!black}{GrowingSleepingMarkovHedge}}

\newcommand{\aggmarkov}{\mhedge}

\jmlryear{2017}
\jmlrworkshop{Algorithmic Learning Theory 2017}
\editors{Steve Hanneke and Lev Reyzin}
\jmlrvolume{76}

\title{Efficient tracking of a growing number of experts}
\author{\name Jaouad Mourtada \email jaouad.mourtada@polytechnique.edu \\
       \addr Centre de Math\'ematiques Appliqu\'ees\\
       \'Ecole Polytechnique\\
       91128 Palaiseau, France
       \AND
       \name Odalric-Ambrym Maillard \email odalric.maillard@inria.fr\\
       \addr Inria Lille - Nord Europe\\       
       59650 Villeneuve d'Ascq, France
     }

 \author{\Name{Jaouad Mourtada} \Email{jaouad.mourtada@polytechnique.edu}\\
 \addr Centre de Math\'ematiques Appliqu\'ees\\
       \'Ecole Polytechnique\\
       91128 Palaiseau, France
 \AND
 \Name{Odalric-Ambrym Maillard} \Email{odalric.maillard@inria.fr}\\
 \addr Inria Lille - Nord Europe\\       
       59650 Villeneuve d'Ascq, France
 }

\begin{document}

\maketitle

\begin{abstract}
We consider a variation on the problem of prediction with expert advice, where new forecasters that were unknown until then may appear at each round.
As often in prediction with expert advice, designing an algorithm that achieves near-optimal regret guarantees is straightforward, using aggregation of experts.
However, when the comparison class is sufficiently rich, for instance when the best expert and the set of experts itself changes over time, such strategies naively require to maintain a prohibitive number of weights (typically exponential with the time horizon).
By contrast, designing strategies that both achieve a near-optimal regret and maintain a reasonable number of weights is highly non-trivial.
We consider three increasingly challenging objectives (simple regret, shifting regret and sparse shifting regret) that extend existing notions defined for a fixed expert ensemble;
in each case, we design strategies that achieve tight regret bounds, adaptive to the parameters of the comparison class, while being computationally inexpensive.
Moreover, our algorithms are anytime, agnostic to the number of incoming experts and completely parameter-free.
Such remarkable results are made possible thanks to two simple but highly effective recipes: first the ``abstention trick'' that comes from the \emph{specialist} framework and enables to handle the least challenging notions of regret, but is limited when addressing more sophisticated objectives.
Second, the ``muting trick'' that we introduce to give more flexibility. We show how to combine these two tricks in order to handle the most challenging class of comparison strategies.
\end{abstract}


\vspace{-3mm}
\section{Introduction}

\vspace{-2mm}
Aggregation of experts is a well-established framework in machine learning \citep{PLG,vovk,gyo99,hauss98}, that provides a sound strategy to combine the forecasts of many different sources. 
This is classically considered in the sequential prediction setting, where at each time step, a learner receives the predictions of experts, uses them to provide his own forecast, and then observes the true value of the signal, which determines his loss and those of the experts. The goal is then to minimize the %worst-case
\emph{regret} of the learner, which is defined as the difference between his cumulated loss and that of the best expert (or combination thereof), no matter what the experts' predictions or the values of the signal are. 

A standard assumption in the existing literature is that the set of experts is known before the beginning of the game. In many situations, however, it is desirable to add more and more forecasters over time.
For instance, in a non-stationary setting one could add new experts trained on a fraction of the signal, possibly combined with change point detection. Even in a stationary setting, a growing number of increasingly complex models enables to account for increasingly subtle properties of the signal without having to include them from the start, which can be needlessly costly computationally (as complex models, which take more time to fit, are not helpful in the first rounds) or even intractable in the case of an infinite number of models with no closed form expression.
Additionally, in many realistic situations some completely novel experts may appear in an unpredicted way (possibly due to innovation, the discovery of better algorithms or the availability of new data), and one would want a way to safely incorporate them to the aggregation procedure. 

In this paper, we study how to amend aggregation of experts strategies in order to incorporate novel experts that may be added on the fly at any time step. Importantly, since we do not know in advance when new experts are made available, we put a strong emphasis on \textit{anytime} strategies, that do not assume the time horizon is finite and known. Likewise, our algorithms should be agnostic to the total number of experts available at a given time.
Three notions of regret of increasing complexity will be defined for growing expert sets, that extend existing notions to a growing expert set. Besides comparing against the best expert, it is natural in a growing experts setting to track the best expert; furthermore, when the number of experts gets large, it becomes profitable to track the best expert in a small pool of good experts.
For each notion, we propose corresponding algorithms with tight regret bounds. 
As is often the case in structured aggregation of experts, the key difficulty  is typically not to derive the regret bounds, but to obtain efficient algorithms. All our methods exhibit minimal time and space requirements that are linear in the number of present experts.
% that only maintain weights for the present experts.

\vspace{-2mm}
\paragraph{Related work.}
This work builds on the setting of prediction with expert advice \citep{PLG,vovk,hw98} that originates from the work on universal prediction \citep{ryabko1984twice, ryabko1988prediction, merhav, gyo99}. We make use of the notion of \textit{specialists} \citep{fss,cv} and its application to \textit{sleeping experts} \citep{kool}, as well as the corresponding standard extensions (Fixed Share,
%% Decreasing Share, Universal Share,
Mixing Past Posteriors) of basic strategies to the problem of \textit{tracking the best expert}~\citep{hw98,kdr13,mpp}; see also~\citet{willems1996coding,shamir1999low} for related work in the context of lossless compression. 
%% As we focus on countable sets of experts (although extensions could be considered), we do not consider applications to online optimization \citep{hazan07,gaill} when e.g. the gradient of the loss is made available.
 Note that, due to its versatility, aggregation of experts has been adapted successfully to a number of applications \citep{monteleoni2011tracking,mcquade2012global, stoltz}.
 It should be noted that
 the literature on prediction with expert advice is split in two categories: the first one focuses on exp-concave loss functions, whereas the second studies convex bounded losses.
 While our work belongs to the first category, it should be possible to transport our regret bounds to the convex bounded case by using time-varying learning rates, as done e.g. by~\citet{hazan09} and~\citet{gyorgy2012efficient}.
% and while it is possible to transport our regret bounds to the convex bounded case, the resulting algorithms are no longer parameter-free since the learning rate now has to be tuned optimally.
 In this case, the growing body of work on the automatic tuning of the learning rate \citep{ftl,llr} as well as alternative aggregation schemes
 \citep{wintenberger2017boa,squint,adanormalhedge} 
% might be of particular interest.
might open the path for even further improvements.

 The use of a growing expert ensemble was already proposed by~\cite{gyo99} in the context of sequentially predicting an ergodic stationary time series, where new higher order Markov experts were introduced at exponentially increasing times (and the weights were reset as uniform); since consistency was the core focus of the paper, this simple ``doubling trick'' could be used, something we cannot afford when new experts arrive more regularly.
 Closer to our approach,  growing expert ensembles have been considered in contexts where the underlying signal may be non-stationary, see e.g. \cite{hazan09,shalizi2011adapting}.
 Of special interest to our problem is~\cite{shalizi2011adapting}, which considers the particular case when one new expert is introduced every $\tau$ time steps, and propose a variant of the Fixed Share (FS) algorithm analogous to our \gmhedge\ algorithm.
 However, their algorithms depend on parameters which have to be tuned depending on the parameters of the comparison class, whereas our algorithms are parameter-free and do not assume the prior knowledge of the comparison class.
 Moreover, we introduce several other algorithms tailored to different notions of regret; in particular, we address the problem of comparing to sequences of experts that alternate between a small number of experts, a refinement that is crucial when the total set of experts grows, and has not been obtained previously in this context.

Another related setting is that of ``branching experts'' considered by~\citet{gofer2013branching}, where each incumbent expert is split into several experts that may diverge later on. Their results include a regret bound in terms of the number of \emph{leading experts} (whose cumulated loss was minimal at some point). Our approach differs in that it does not assume such a tree-like structure: a new entering forecaster is not assumed to be associated to an incumbent expert. More importantly, while~\cite{gofer2013branching} compare to the leaders in terms of cumulated loss (since the beginning of the game), our methods compete instead with sequences of experts that perform well on some periods, but can predict arbitrarily bad on others; this is harder, since the loss of the optimal sequence of experts can be significantly smaller than that of the best expert.  

%% Crucially, the notion of "sparse experts" considered in their work differs significantly from ours. They consider the case of few "leaders" in terms of *cumulated regret* (since the beginning of the game). Our methods compete instead with the best partition between a small number of experts that can perform well on some time intervals and arbitrarily bad on others, which is harder: the loss of the optimal sparse sequence of experts can be significantly smaller than the cumulated loss of the best expert.

\vspace{-2mm}
\paragraph{Outline.}
Our paper is organized as follows. After introducing the setting, notations and the different comparison classes, we provide in Section~\ref{sec:overview} an overview of our results, stated in less general but more directly interpretable forms. Then, Section~\ref{sec:exp} introduces the exponential weights algorithm and its regret, a classical preliminary result that will be used throughout the text. Sections~\ref{sec:grow-spec},~\ref{sec:grow-seq} and~\ref{sec:grow-sleep} form the core of this paper, and have %present
the same structure: a generic result is first stated in the case of a fixed set of experts, before being turned into a strategy in the growing experts framework. Section~\ref{sec:grow-spec} starts with the related \emph{specialist} setting and adapts the algorithm into an anytime growing experts algorithm, with a more general formulation and regret bound involving \emph{unnormalized priors}. Section~\ref{sec:grow-seq} proposes an alternative approach, which casts the growing experts problem as one of competing against \emph{sequences} of experts; this approach proves more flexible and general for our task, but perhaps surprisingly we can also recover algorithms that are essentially equivalent to the aggregation of growing experts with an unnormalized prior. Finally, the two approaches are combined in Section~\ref{sec:grow-sleep} in the context of \emph{sleeping experts}, where we reinterpret the algorithm of~\cite{kool} and extend it to more general priors before adapting it to the growing experts setting.

\vspace{-2mm}
\section{Overview of the results}
\label{sec:overview}

\vspace{-1mm}
Our work is framed in the classical setting of \emph{prediction with expert advice}~\citep{vovk,PLG}, which we adapt to account for a growing number of experts. The problem is characterized by its \emph{loss function} $\ell : \X \times \Y \to \R$, where $\X$ is a convex \emph{prediction space}, and $\Y$ is the \emph{signal space}.

Let $M_t$ be the total number of experts at time $t$, and $m_t = M_t-M_{t-1}$ be the number of experts introduced at time $t$. We index experts by their entry order, so that expert $i$ is the $i^{th}$ introduced expert and denote $\tau_i = \min\{ t\geq 1 : i \leq  M_{t} \}$ its \textit{entry time} (the time at which it is introduced). We say we are in the \emph{fixed expert set} case  when $M_t=M$ for every $t\geq 1$ and in the \emph{growing experts setting} otherwise.
At each  step $t\geq 1$, the experts $i = 1, \dots, M_t$ output their predictions $x_{i,t} \in \X$, which the learner uses to build $x_t \in \X$; then, the environment decides the value of the signal $y_t \in \Y$, which sets the losses $\ell_t = \ell (x_t,y_t)$ of the learner and  $\ell_{i,t} = \ell (x_{i,t} , y_t)$ of the experts.

\vspace{-1mm}
\paragraph{Notations.}

Let $\pr_M$ be the \emph{probability simplex}, \ie the set of probability measures over the set of experts $\{ 1, \dots, M \}$.
We denote by $\kll \cdot \cdot$ the \emph{Kullback-Leibler divergence}, defined for $\bm u, \bm v \in \pr_M$ by 
$
\kll {\bm u} {\bm v} = \sum_{i=1}^M u_i \ln \frac{u_i}{v_i}
  \geq
  \!
\ 0
$.
%   \footnote{Indeed, one has by concavity of the $\log$ function: $\kll {\bm u}{\bm v}
%   = - \sum_{i=1}^M u_i \ln \frac{v_i}{u_i}
%   \geq - \ln \sum_{i=1}^M u_i \frac{v_i}{u_i}
%   % = -\ln \sum_{i=1}^M v_i
% %  = - \ln 1
%   =0$.}
%\end{equation}

\vspace{-1mm}
\paragraph{Loss function.}

Throughout this text, we make the following standard assumption\footnote{This could be readily replaced (up to some cosmetic changes in the statements and their proofs) by the more general \emph{$\eta$-mixability} condition~\citep{vovk}, that allows to use higher learning rates $\eta$ for some loss functions (such as the square loss, but {not} the logarithmic loss) by using more sophisticated combination functions.
  %However, we restrict to exp-concave losses in our text for convenience, since using alternative combination function would call for cosmetic changes in the statements of the results and their proofs.
} on the loss function~\citep{PLG}. 

\vspace{-3mm}
\begin{assumption}\label{ass:expconcave} The loss function $\ell$ is \emph{$\eta$-exp-concave} for some $\eta >0$, in the sense that $\exp (- \eta \, \ell(\cdot, y))$ is concave on $\X$ for every observation $y\in \Y$.
  This is % readily seen to be
  equivalent to the inequality
\begin{equation}
  \label{eq:exp-conc}
  \ell \left(\sum_{i=1}^M v_{i} \, x_{i} , y \right)
  \leq - \frac 1\eta \ln \sum_{i=1}^M v_i \, e^{-\eta\, \ell(x_i,y)}
\end{equation}

\vspace{-2mm}\noindent
for every $y \in \Y$, $\bm x = (x_i)_{1\leq i \leq M} \in \X^M$ and $\bm v = (v_i)_{1\leq i \leq M} \in \pr_M$.
%% $\pr_M$ is defined later...
\end{assumption}

\vspace{-6mm}
\begin{rem}
  An important example in the case when $\X$ is the set of probability measures over $\Y$ is the \emph{logarithmic} or \emph{self-information} loss $\ell(x,y) = -\log x( \{ y \} )$ for which the inequality holds with $\eta=1$, and is actually an equality. Another example of special interest is the quadratic loss on a bounded interval: 
  indeed, for $\X = \Y = [a,b] \subset \R$, $\ell(x,y) = (x-y)^2$ is $\frac 1{2 (b-a)^2}$-exp-concave.
  % \footnote{In this case, however, a faster learning rate $\eta = \frac{2}{(b-a)^2}$ can be achieved by using another combination function than the convex combination, see Appendix~\ref{ap:mix} for more details.}.
\end{rem}

%% Alternative : we will see in section how to extend...
%%% All the bounds proven in this setting transfer\footnote{With an additional term linear in $\eta$, which leads to issues of optimal tuning of this parameter.} to the case

\vspace{-2mm}

Several notions of regret can be considered in the growing expert setting. 
We review here three of them, each corresponding to a specific comparison class; we show the kind of bounds that our algorithms achieve, to illustrate the more general results stated in the subsequent sections.
We provide more uniform bounds in Appendix~\ref{ap:info}, and compare them with information-theoretic bounds.
% their optimality is discussed

\vspace{-1mm}
\paragraph{Constant experts.}

 Since the experts only output predictions after their entry time, it is natural to consider the \emph{regret} with respect to each expert $i\geq 1$ over its time of activity, namely the quantity
 
  \vspace{-5mm}
\begin{equation}
  \label{eq:reg-const}
  \sum_{t = \tau_i}^T (\ell_t - \ell_{i,t})
\end{equation}

\vspace{-2mm}\noindent
for every $T \geq \tau_i$. Note that this is equivalent to controlling~(\ref{eq:reg-const}) for every $T \geq 1$ and $i \leq M_T$. Algorithm~\ghedge\ is particularly relevant in this context; with the choice of (unnormalized) prior weights %
%% \footnote{In fact, this bound can be slightly refined in the case when new experts do not enter at each round, see the discussion in Section~\ref{sub:ghedge}.}
$\pi_i = \frac 1{\tau_im_{\tau_i}}$, it achieves the following regret bound: for every $T\geq 1$ and $i \leq M_T$,

\vspace{-4mm}
\begin{equation}
  \label{eq:reg-constb}
  \sum_{t=\tau_i}^T (\ell_t - \ell_{i,t}) \leq
  \frac 1\eta \log m_{\tau_i} + \frac 1\eta \log \tau_i
  + \frac 1\eta \log (1+ \log T) \, .
\end{equation}

  \vspace{-1mm}
\noindent 
This bound has the merit of being simple, virtually independent of $T$ 
%% (as the $\log \log T$ term is bounded by $4$ in any practical situation)
and independent of the number of experts $(m_t)_{t>\tau_i}$ added after $i$. Several other instantiations
of the general regret bound of \ghedge\ (Theorem~\ref{thm:ghedge}) are given in Section~\ref{sub:ghedge}.

\vspace{-2mm}
\paragraph{Sequences of experts.}

Another way to study growing expert sets is to view them through the lens of sequences of experts. Given a sequence of experts $i^T = (i_1, \dots, i_T)$, we measure the performance of a learning algorithm against it in terms of the \textit{cumulative regret}:

\vspace{-5mm}
\begin{equation}
L_T - L_T(i^T) = \sum_{t=1}^T \ell_t - \sum_{t=1}^T \ell_{i_t,t}\,,
\end{equation}
% Since it is provably impossible to get a controlled regret against an \textit{arbitrary} sequence $i^\infty$ in full generality, we consider in the sequel different constraints on the sequence of experts that lead to different notions of regret, from the least to the most challenging one:

\vspace{-2mm}
\noindent 
In order to derive meaningful regret bounds, some constraints have to be imposed on the comparison sequence; hence, we consider in the sequel different types of comparison classes that lead to different notions of regret, from the least to the most challenging one:

\medskip

\vspace{-1mm}
\noindent{\bf (a) Sequences of fresh experts.}
These are \emph{admissible} sequences of experts $i^T$, in the sense that $i_t \leq M_t$ for $1 \leq t\leq T$ (so that $\ell_{i_t,t}$ is always well-defined) that only switch to \emph{fresh} (newly entered) experts, \ie if $i_t\neq i_{t-1}$, then $M_{t-1}+1 \leq i_t \leq M_t$. More precisely, for each $\bm \sigma = (\sigma_1 , \dots, \sigma_k)$ with $1 < \sigma_1 < \dots < \sigma_k \leq T$, $\fseq_T (\bm \sigma)$ denotes the set of sequences of fresh  experts whose only shifts occur at times $\sigma_1, \dots, \sigma_k$.
Both the switch times $\bm \sigma$ and the number of shifts $k$ are assumed to be unknown, although to obtain controlled regret one typically needs $k \ll T$.
Comparing to sequences of fresh experts is essentially equivalent to comparing against constant experts; algorithms \ghedge\ and \fmhedge\ with
%%%% $\pi_i = \frac 1{\tau_i m_{\tau_i}}$
$\pi_i = \frac 1{m_{\tau_i}}$ %%%%
achieve, for \emph{every} $T \geq 1$, $k \leq T-1$ and %$1  =: \sigma_0    < \sigma_1 < \dots < \sigma_k \leq T$
  %% modify for fmhedge
  $\bm \sigma = (\sigma_j)_{1\leq j \leq k}$
    (Theorems~\ref{thm:ghedge} and~\ref{thm:fmhedge}):
  \begin{equation}
    \label{eq:reg-fseq}
    L_T - \inf_{i^T \in \fseq_T (\bm \sigma)} L_T (i^T)
    \leq
%    \frac 1\eta
%    \left\{
%      \log m_1 +  \sum_{j=1}^k (\log m_{\sigma_j} + \log \sigma_j + \log (1+ \log \sigma_j)) +  \log (1 + \log T)
%    \right\}
    \frac 1\eta
    \bigg\{
      \log m_1 +  \sum_{j=1}^k (\log m_{\sigma_j} + \log \sigma_j) +
      %%%% (k+1) \log (1 + \log T)
      \log T %%%%
    \bigg\}
  \end{equation}
  In particular, the regret with respect to any sequence of fresh experts with $k$ shifts is bounded by $\frac 1\eta \left( (k+1) \log \max_{1\leq t \leq T} m_t + 
    %%%% k \log T + (k+1) \log (1+\log T)
    (k+1) \log T %%%%
  \right)$.

\medskip    
\noindent {\bf (b) Arbitrary admissible sequences of experts.}
  Like before, these are admissible sequences of experts that are piecewise constant with a typically small number of shifts $k$, except that  shifts to \emph{incumbent} (previously introduced) experts $i_t \leq M_{t-1}$ are now authorized. Specifically, given $\bm \sigma^0 = (\sigma^0_1 , \dots, \sigma^0_{k_0})$ and $\bm\sigma^1 = (\sigma^1_1 , \dots, \sigma^1_{k_1})$, we denote by $\aseq_T (\bm \sigma^0; \bm \sigma^1)$ the class of admissible sequences whose switches to fresh (resp. incumbent) experts occur only at times $\sigma^0_1 < \dots < \sigma^0_{k_0}$ (resp. $\sigma^1_1 < \dots < \sigma^1_{k_1}$). 
  By Theorem~\ref{thm:gmhedge}, algorithm \gmhedge\ with
  %%%% $\pi_i = \frac{1}{\tau_i m_{\tau_i}}$
  $\pi_i = \frac{1}{m_{\tau_i}}$ %%%%
  and $\alpha_t = \frac 1t$ satisfies, for \emph{every} $T \geq 1$, $k_0,k_1$ with $k_0+k_1 \leq T-1$ and $\bm \sigma^0, \bm \sigma^1$:
  \begin{equation}    
    \label{eq:reg-aseq}    
%    \begin{split}      
  L_T - \inf_{i^T \in \aseq_T (\bm \sigma^0 ; \bm \sigma^1)} L_T (i^T)
  \leq % \\
   \frac 1\eta
    \bigg\{
      \log m_1 +  \sum_{j=1}^k (\log m_{\sigma_j} + \log \sigma_j)
      + \sum_{j=1}^{k_1} \log \sigma^1_j
      %%%% + \log T + k \log (1 + \log T)
      + 2 \log T %%%%      
    \bigg\}
%  \end{split}  
\end{equation}
where $k=k_0 + k_1$ and $\sigma_1 < \dots < \sigma_k$ denote \emph{all} shifts (either in $\bm \sigma^0$ or in $\bm \sigma^1$). Note that the upper bound~\eqref{eq:reg-aseq} may be further relaxed as $\frac 1\eta \left( (k+1) \log \max_{1\leq t \leq T} m_t + (k_0 + 2 k_1 +
  %%%% 1
  2 %%%%
  ) \log T
  % + k \log (1+\log T)
\right) $%%%% (omitting the negligible doubly logarithmic term)
.

  \medskip    
\noindent {\bf (c) Sparse sequences of experts.}
 These are admissible sequences $i^T$ of experts that are additionally \emph{sparse}, in the sense that they alternate between a small number $n \ll M_T$ of experts; again, $n$ may be unknown in advance.
  %% and n << k
  Denoting
%  $\sseq_T (\bm \sigma, n)$ the class of sequences with shifts in $\bm \sigma$ and taking at most $n$ values,
  $\sseq_T (\bm \sigma, E)$ the class of sequences with shifts in $\bm \sigma$ and taking values in the subset of experts $E = \{ e_1, \dots, e_n \}$,
 algorithm~\gsmhedge\ with $\pi_i = \frac{1}{\tau_i m_{\tau_i}}$ and $\alpha_t = \beta_t =\frac 1t$ achieves, for every $T \geq 1$, $E \subset \{ 1, \dots, M_T \}$ and $\bm \sigma$,
  \begin{equation}
    \label{eq:reg-last}
          L_T -  \inf_{i^T \in \sseq_T (\bm \sigma, E)} L_T (i^T)
          \leq
       \frac 1\eta \sum_{p=1}^n \big(\ln \tau_{e_p} + \ln \frac {m_{\tau_{e_p}}}n \big)
       + \frac 1\eta n \log (2 T)
       + \frac 2\eta \sum_{j=1}^k  \log \sigma_j.
  \end{equation}
 In particular, the regret with respect to every admissible sequence of $T$ experts with at most $k$ shifts and taking at most $n$ values is bounded by $\frac 1\eta \left( n \log \frac{\max_{1\leq t \leq T} m_t}{n} + 2 n \log (\sqrt{2} T) + 2 k \log T \right)$.

%% \subsection{Properties 
%%   %Requirements
%%   of our algorithms}
%% \label{sub:alg}
%%
%% Anytime; agnostic to the future numbers of incoming experts $m_t$; parameter-free; $O(M_t)$ time at step $t$; regret bounds adaptive to the parameters of the comparison class.

 The main results of this text are Theorem~\ref{thm:gmhedge}, a powerful parameter-free generalization of~\cite[Theorem~2]{shalizi2011adapting}, and Theorem~\ref{thm:gsmhedge}, which adapts results of~\cite{mpp,kool} to sequentially incoming forecasters, and has no precedent in this context.
 
 %%% We provide uniform bounds (that are more directly interpretable than the adaptive bounds presented here) a comparison with information-theoretic bounds in Appendix

\vspace{-3mm}
\section{Preliminary: the exponential weights algorithm}
\label{sec:exp}

\vspace{-1mm}
First, we introduce the simple but fundamental \emph{exponential weights}
or \emph{Hedge algorithm} \citep{vovk,PLG}, designed to control the regret  $ L_T - L_{i,T} = \sum_{t=1}^T \ell_t - \sum_{t=1}^T \ell_{i,t}$ for a fixed set of experts $\{ 1, \dots, M\}$.
The algorithm depends on a \emph{prior distribution} $\bm \pi \in \pr_M$ on the experts and predicts~as
\begin{equation}
  \label{eq:ewa}  
  x_{t}
  = \frac{\sum_{i=1}^M w_{i,t} \, x_{i,t}}{\sum_{i=1}^M w_{i,t}} \quad\text{with}\quad w_{i,t} = \pi_i \, e^{-\eta L_{i,t-1}} \, .
\end{equation}
Equivalently, it forecasts $x_t = \sum_{i=1}^M v_{i,t} \, x_{i,t}$, where the weights $\bm v_t \in \pr_M$ are sequentially updated in the following way: $\bm v_1 = \bm\pi$ and,
after each round $t\geq 1$, $\bm v_{t+1}$ is set to the \emph{posterior} distribution $\bm v_t^m$ of $\bm v_t$ given the losses $(\ell_{i,t})_{1\leq i \leq M}$, defined by
\begin{equation}
  \label{eq:post}
  v_{i,t}^m = \frac{v_{i,t}\, e^{-\eta \, \ell_{i,t}}}
  {\sum_{j=1}^M v_{j,t}\, e^{-\eta \, \ell_{j,t}}} \, .
\end{equation}
% The use of the terms \emph{prior} and \emph{posterior} stems from the observation that, in the case of the logarithmic loss and for $\eta =1$, the exponential weights algorithm reduces to Bayesian averaging of the experts (seen as probability distributions on the signal).
All  subsequent regret bounds will rely on the following standard regret bound (see Appendix~\ref{ap:prop1}), by reducing complex forecasting strategies to the aggregation of experts under a suitable prior.

%% The following well-known regret bound, whose proof we provide in Appendix~\ref{ap:prop1}, will be used repeatedly throughout this article:

\vspace{-1mm}
\begin{proposition}
  [{\citet[Corollary 3.1% p. 47
    ]{PLG}}]
  \label{prop:exp}
%  Assume the loss function $\ell$ is $\eta$-exp-concave.
  Irrespective of the values of the signal and the experts' predictions, the exponential weights algorithm~(\ref{eq:ewa}) with
  % learning rate $\eta$ and
  prior $\bm\pi$ achieves
\vspace{-1mm}
  \begin{equation}
    \label{eq:reg-exp}
    L_T - L_{i,T} \leq \frac 1\eta \log \frac 1{\pi_i}
  \end{equation}    
  for each $i=1, \dots , M$ and $T \geq 1$. 
  More generally, for each probability vector $\bm u \in \pr_M$, % we have

\vspace{-5mm}
  \begin{equation}
    \label{eq:reg-comb}
    L_T - \sum_{i=1}^M u_i\, L_{i,T}
    \leq \frac 1\eta \, \kll {\bm u} {\bm \pi} \, .
  \end{equation}
\end{proposition}
\vspace{-3mm}
Choosing a uniform prior $\bm \pi = \frac 1M \bm 1$ yields an at most $\frac 1\eta \log M$ regret with respect to the best expert.
%%  When the number of experts is countably infinite, one can still get useful regret bounds for arbitrary experts predictions, by choosing slowly decreasing normalized weights $\pi_i$, %$i \in \N^*$

 \vspace{-1mm}
\section{Growing experts and specialists: the ``abstention trick''}
\label{sec:grow-spec}

\vspace{-1mm}
A natural idea to tackle the problem of a growing number of experts is to cast it in the related setting of \emph{specialists}, introduced by~\citet{fss}.
We present the specialist setting and the related ``specialist trick'' identified by~\cite{cv} (which we will call the ``abstention trick''), which enables to convert any expert aggregation algorithm into a specialist aggregation algorithm.
These ideas are then applied to the growing expert ensemble setting, which allows us to control the regret with respect to \emph{constant} experts of equation~(\ref{eq:reg-const}); a refinement is introduced along the way, the use of \emph{unnormalized priors}, that gives more flexibility to the algorithm and its regret bounds. 

\vspace{-1mm}
\subsection{Specialists and their aggregation}
%% The specialist trick/setting
\label{sub:spec}

\vspace{-1mm}
In the specialist setting, we have access to \emph{specialists}
%%%%\footnote{In the literature, specialists are sometimes called ``sleeping experts''. In this paper, we refrain from using this synonym, since we use the term ``sleeping experts'' to refer to a slightly different concept (see Section~\ref{sub:sleep}).}
$i \in \{ 1, \dots, M \}$ that only output predictions at certain steps, while refraining from predicting the rest of the time.
In other words, at each step $t\geq 1$, only a subset $A_t \subset \{ 1 , \dots , M\}$ of \emph{active} experts output a prediction $x_{i,t} \in \X$.

In order to adapt any expert aggregation strategy to the specialists setting, 
a crucial idea due to \citet{cv} is to ``complete'' the specialists' predictions by attributing to inactive specialists $i \not\in A_t$ a forecast equal to that of the aggregating algorithm.
Although this seems circular, it can be made precise by observing that the only way to simultaneously satisfy the conditions

\vspace{-5mm}
\begin{equation}
  \label{eq:conditions}
  x_{t} = \sum_{i=1}^M v_{i,t}\, x_{i,t}
  \quad \mbox{ and } \quad
  x_{i,t} = x_t \
  \mbox{ for any } i \not \in A_t
\end{equation}
is to take
% $x_{t} = x_{i,t} = \frac{\sum_{i\in A_t} v_{i,t}\, x_{i,t}}{\sum_{i\in A_t} v_{i,t}}$ for $i \not\in A_t$.

\vspace{-8mm}
\noindent 
\begin{equation}
    \label{eq:extension}
    x_{t} = x_{i,t} = \frac{\sum_{i\in A_t} v_{i,t}\, x_{i,t}}{\sum_{i\in A_t} v_{i,t}}
    \quad \mbox{ for } i \not\in A_t \, .
  \end{equation}
  We call this technique the ``abstention trick'', since it consists in attributing to inactive specialists a forecast that will not affect the  voting outcome. %, which is equal to the average of the active specialists' predictions. 
  In the case of the exponential weights algorithm, this leads to the \emph{specialist aggregation} algorithm with prior $\bm \pi$, which forecasts
  \begin{equation}    
    \label{eq:shedge}    
    x_t = \frac{\sum_{i\in A_t} w_{i,t} \, x_{i,t}}    
    {\sum_{i\in A_t} w_{i,t}}
    \quad \mbox{with} \quad
    w_{i,t} = \pi_i \, e^{-\eta L_{i,t-1}} \, ,
  \end{equation}
where we denote, for each specialist $i$ and $t\geq 1$, $L_{i,t} := \sum_{s\leq t \pp i \in A_s} \ell_{i,s} + \sum_{s \leq t \pp i \not\in A_s} \ell_s$.

\vspace{-2mm}
\begin{rem}\label{rem:equality}
  The exp-concavity inequality $e^{-\eta\, \ell_t} \geq \sum_{i=1}^M v_{i,t} \, e^{-\eta \,\ell_{i,t}}$ shows that $v_{i,t+1} \geq v_{i,t}$ for any $i \not\in A_t$.
  In the case of the logarithmic loss, for $\eta =1$ this inequality becomes an equality, thus the weights of inactive specialists remain unchanged: $v_{i,t+1} = v_{i,t}$.
\end{rem}
% \noindent
\vspace{-2mm}
Since the specialist aggregation consists of the exponential weights on the 
% the exponentially weighted aggregation of the experts defined by the extension of
extended predictions~(\ref{eq:extension}), and since for this extension one has
    $\sum_{t=1}^T (\ell_t - \ell_{i,t})
    =  \sum_{t\leq T \pp i\in A_t} (\ell_{t} - \ell_{i,t})$, 
 Proposition~\ref{prop:exp} implies:

 \vspace{-2mm}
\begin{proposition}[{\citet[Theorem 1]{fss}}]
  \label{prop:spec}
  The specialist aggregation with prior $\bm \pi$ achieves the following regret bound: for each specialist $i$ and every $T \geq 1$,
  \begin{equation}
    \label{eq:reg-spec1}
    \sum_{t\leq T \pp i\in A_t} (\ell_{t} - \ell_{i,t})
    \leq \frac 1\eta \log \frac 1{\pi_i}  \, .
  \end{equation}

  \vspace{-4mm}
  \noindent 
  Moreover, for each probability vector $\bm u \in \pr_M$,\quad 
$    \displaystyle{
    \sum_{i=1}^M u_i \sum_{t\leq T \pp i\in A_t} (\ell_{t} - \ell_{i,t})
    \leq \frac 1\eta \, \kll {\bm u} {\bm \pi} \, .}$
\end{proposition}

\begin{rem}
  Note that the sets $A_t$ of active specialists do not need to be known in advance.
  % for Strategy~\ref{alg:spec} to be implemented/applied.
\end{rem}

\vspace{-4mm}
\subsection{Adaptation to growing expert ensembles: \ghedge}
%\subsection{Growing experts in the specialist setting: strengths and limitations}
\label{sub:ghedge}

\vspace{-1mm}
Growing experts can naturally be seen as specialists, by setting $A_t := \{ 1, \dots, M_t\}$; moreover, through this equivalence, the quantity controlled by Proposition~\ref{prop:spec} is precisely the regret~(\ref{eq:reg-const}) with respect to \emph{constant experts}. In order to apply the results on specialist aggregation to the growing expert setting, it remains to specify exactly which total set of specialists is considered.

\paragraph{Fixed time horizon.}
In the simplest case when both the time horizon $T$ and the eventual number of experts $M_T$ are known, the eventual set of experts (at time $T$) is known, and we can take the finite specialist set to be $\{ 1, \dots, M_T\}$. Therefore, given any probability vector $\bm \pi = (\pi_1, \dots, \pi_{M_T})$, we can use the aggregation of specialists, with the regret bound~(\ref{eq:reg-spec1}).
In particular, the choice of $\pi_i = \frac 1{M_T}$ for $i =1 , \dots, M_T$ yields the uniform regret bound $\frac 1\eta \log M_T$.

\vspace{-1mm}
\paragraph{Anytime algorithm, normalized prior.}

The fixed horizon approach is somewhat unsatisfactory, since we are typically interested in algorithms that are anytime and agnostic to $M_t$. 
To achieve this goal, a better choice is to take the infinite set of specialists $\N^*$. 
Crucially, the aggregation of this infinite number of specialists can be implemented in finite time, by introducing the weight of an expert \textit{only when it enters}.
Given a probability vector $\bm\pi = (\pi_i)_{i\geq 1}$ on $\N^*$, this leads to the anytime strategy \ghedge\ described below. A straightforward adaptation of Propositions~\ref{prop:exp} and~\ref{prop:spec} to a countably infinite set of experts shows that this strategy achieves, now for \emph{every} $T \geq 1$ and $i \leq M_T$, the regret bound~(\ref{eq:reg-spec1}).
However, %%in this case
we are constrained by the fact that $\bm \pi$ must be a probability %%%measure
on $\N^*$.%
% it turns out that
% this assumption can be removed completely, leading to tighter bounds and more flexible strategies.

% A limitation of this result is that we cannot get the ``uniform algorithm'' with the same regret bound $\log M_t$ for each expert.

\vspace{-1mm}
\paragraph{Anytime algorithm, unnormalized prior.}

We now turn to the most general analysis, which subsumes and improves the previous two. Now, we let $\bm \pi = (\pi_i)_{i \geq 1}$ denote a sequence of \emph{arbitrary} positive weights, that are no longer assumed to sum to $1$. These weights do not need to be set in advance: the weight $\pi_i$ can be chosen when expert $i$ enters, so that at this step $\tau_i$, $(m_{t})_{t\leq \tau_i}$ and $(M_{t})_{t\leq \tau_i}$ are known, even if they were unknown at the beginning; in particular, $\pi_i$ may depend on these quantities. We now consider the anytime algorithm~\ghedge.%

%% A key observation is that one can in fact use Algorithm~\ref{alg:ghedge} with arbitrary, \emph{unnormalized} weights $\pi_i >0$. This observation gives more flexibility in the design of the aggregation strategy, as well as a refined regret bound.

  \begin{algorithm}            
  \caption{\ghedge\ --- Anytime aggregation of growing experts}
\label{alg:ghedge}              
\begin{algorithmic}[1]
  \STATE \textbf{Parameters:} Learning rate $\eta >0$, weights on the experts $\bm \pi = (\pi_i)_{i \geq 1}$.
  \STATE \textbf{Initialization:} Set $w_{i,1} = \pi_i$ for $i=1, \dots, M_1$.
  \FOR{$t = 1, 2, \dots$} 
  \STATE Receive predictions $(x_{1,1}, \dots, x_{M_t,t})\in \X^{M_t}$ from the experts, and predict
  \vspace{-4pt}
  \begin{equation}
    \label{eq:pred-grow1}
    x_t = \frac{\sum_{i=1}^{M_t} w_{i,t} \, x_{i,t}}{\sum_{i=1}^{M_t} w_{i,t}} \, .
  \end{equation}
  %% Important: this implementation only maintains weights for entered experts.
  
  \vspace{-1mm}
  \STATE Observe $y_t \in \Y$, and derive the losses $\ell_t = \ell (x_t, y_t)$ and $\ell_{i,t} = \ell (x_{i,t}, y_t)$.
  \STATE Update the weights by $w_{i,t+1} = w_{i,t} \, e^{-\eta\, \ell_{i,t}}$ for $i = 1, \dots, M_t$. Moreover, introduce the weights $w_{i,t+1} = \pi_i \, e^{-\eta \, L_{t}}$ for $M_t+1 \leq i \leq M_{t+1}$.
  \ENDFOR
\end{algorithmic}
\end{algorithm}

\vspace{-2mm}
\begin{theorem}
  \label{thm:ghedge}
  Let $\bm \pi = (\pi_i)_{i \geq 1}$ be an arbitrary sequence of positive weights.
  Then, algorithm \ghedge\ achieves the following regret bound: for every $T \geq 1$ and $i \leq M_T$, 

\vspace{-5mm}
  \begin{equation}
    \label{eq:reg-ghedge}
  \sum_{t=\tau_i}^T (\ell_t - \ell_{i,t})
  \leq \frac 1\eta \log \bigg( \frac 1{\pi_i} \sum_{j=1}^{M_T} \pi_j \bigg) \, .
  \end{equation}
  
  \vspace{-1mm}
  \noindent
  Additionally, its time and space complexity at each step $t\geq 1$ is $O(M_t)$.
%  
%  More generally, assume we are given a set $\M$ of specialists, as well as a positive weight function $\pi : \M \to \R_+^*$. Assume that, at each time step $t\geq 1$, the set $A_t$ of active specialists is finite. Then, denoting $A_{\leq t} = \bigcup_{1\leq s \leq t} A_s$, the aggregation of specialists\footnote{Denoting, for each specialist $i$ and $t\geq 1$, $L_{i,t} = \sum_{s\leq t \pp i \in A_s} \ell_{i,s} + \sum_{s \leq t \pp i \not\in A_s} \ell_s$.}
%    \[ x_t = \frac{\sum_{i \in A_t} \pi (i) \, e^{-\eta\, L_{i,t-1}} \, x_{i,t}}
%      {\sum_{i \in A_t} \pi (i) \, e^{-\eta\, L_{i,t-1}}}
%    \]
%    achieves the following regret bound: for each $T\geq 1$ and $i \in \M$, we have
%    \[
%      \sum_{t\leq T \pp i \in A_t} (\ell_{t}- \ell_{i,t})
%      \leq \frac 1{\eta} \log
%      \bigg( \frac{1}{\pi (i)} \sum_{j \in A_{\leq T}} \pi (j)  \bigg) \, .
%    \]
%    \ja{Denote $\Pw (A) := \sum_{i \in A} \pw (i) $, so that the bound writes  $\frac 1\eta \log \frac{\Pw (A_{\leq T})}{\pw (i)}$, analogous to the previous form ?}
\end{theorem}

\vspace{-2mm}
\noindent
%% Due to space constraints, we postpone
We provide the proof of Theorem~\ref{thm:ghedge} in Appendix~\ref{ap:thm-ghedge}.
Let us now discuss a few choices of priors, with the corresponding regret bounds~(\ref{eq:reg-ghedge}) (omitting the $\frac 1\eta$ factor).

$\bullet$ With $\pi_i = 1$, we get $\log M_T$, but now with an anytime algorithm. Since $\sum_{i=1}^M \frac 1 i \leq 1 + \sum_{i=2}^M \int_{i-1}^i \frac{dx}x = 1 + \log M$, 
  the choice of  $\pi_i = \frac 1i$ yields $\log i + \log (1+\log M_T)$.

 $\bullet$ The above bounds depend on the index $i \geq 1$, and hence arbitrarily distinguish experts entered at the same time. More natural bounds would only depend on the entry time $\tau_i$, which is achievable since $\pi_i$ can be chosen when $i$ enters, and thus depend on $\tau_i$.
%%%$\bullet$ The above bounds are somewhat unsatisfactory, since they depend on the index $i \geq 1$ instead of the entry time $\tau_i$ (and hence arbitrarily distinguish experts entered at the same time). As noted before, $\pi_i$ can be chosen when $i$ enters, and thus can depend on $\tau_i$.
 Setting\footnote{In fact, this can be slightly refined when $m_t =0$ for most steps $t$.
   In this case, denoting for $t \geq 1$: $s(t) = | \{ t' \leq t \mid m_{t'} \geq 1 \} |$, we can take $\pi_i = \frac 1{s (\tau_i) \, m_{\tau_i}}$ and get a regret bound $\frac 1\eta \left\{ \log m_{\tau_i} + \log s (\tau_i) + \log (1+ \log s(T)) \right\}$.}
 $\pi_i = \frac{1}{m_{\tau_i}} \nu_{\tau_i}$, where $\bm \nu = (\nu_t)_{t \geq 1}$ is a positive sequence set in advance, we get
  
\vspace{-6mm}
  \begin{equation}
  \label{eq:reg-1c2}
   \log m_{\tau_i} +  \log \frac 1{\nu_{\tau_i}} +   \log \sum_{t=1}^T \nu_t\,.
\end{equation}

\vspace{-2mm}
\noindent
Amongst the many possible choices for $\nu_t$, one may consider  $\nu_t = 1$ for which~(\ref{eq:reg-1c2})
%the above bound
becomes $\log m_{\tau_i} + \log T$, while $\nu_t = \frac 1t$ yields the improved bound
$\log m_{\tau_i} + \log \tau_i + \log (1+\log T)$. 
 Note that neither choice is summable, and that 
a choice of summable weights (\eg $\nu_t = t^{-\alpha}$, $\alpha>1$ or $\nu_t = \frac{1}{t\log^2(t+1)}$) generally leads to worse or less interpretable bounds. The first choice ($\nu_t = 1$) is simple, while the second one ($\nu_t = 1/t$) trade-offs simplicity and quality of the bound.

$\bullet$ Another option is to set $\pi_i = \upsilon_{\tau_i}$, where $\bm \upsilon = (\upsilon_t)_{t \geq 1}$ is an arbitrary sequence set in advance.
  % (something we could not afford previously, where we had to divide by $m_{\tau_i}$ to ensure normalized weights),
  The bound becomes
  
   \vspace{-10mm}
\begin{equation}
  \label{eq:reg-1c3}
   \log \frac 1{\upsilon_{\tau_i}} +   \log \sum_{t=1}^T m_t \upsilon_t
\end{equation}
which is more regular than the bound~(\ref{eq:reg-1c2})
when $m_t$ alternates between small and large values%
, since it depends on a cumulated quantity instead of just $m_{\tau_i}$. 
For $\upsilon_t = 1$ (\ie $\pi_i =1$) this is just $\log M_T$. Alternatively, for $\upsilon_t = \frac 1t$ this becomes $\log \tau_i + \log \sum_{t=1}^T \frac{m_t}t$.
%%%, where $\sum_{t=1}^T \frac{m_t}t$ can be rewritten $\sum_{t=1}^{T-1} \frac{M_t}{t(t+1)} + \frac{M_T}{T}$.

\paragraph{Regret against sequences of fresh experts.}

\vspace{-1mm}
Theorem~\ref{thm:ghedge} provides a regret bound against any \emph{static} expert, \ie any constant choice of expert, albeit in a growing experts setting. However, this means that the regret is controlled only on the period $\intint{\tau_i}{T}$ when the expert actually emits predictions. An alternative way to state Theorem~\ref{thm:ghedge} is in terms of \emph{sequences of fresh experts}.
Indeed, Theorem~\ref{thm:ghedge} implies that, for every sequence of fresh experts $i^T$ with switching times $\sigma_1 < \dots < \sigma_{k}$ (with the additional conventions $\sigma_0 := 1$ and $\sigma_{k+1} := T+1$), algorithm \ghedge\ achieves:

\vspace{-5mm}
\begin{equation}
  \label{eq:reg-fresh-ghedge}
  %% L_T - \inf_{i^T \in \fseq_T (\bm \sigma)} L_T (i^T)
  L_T - L_T (i^T)
  = \sum_{j=0}^k \sum_{t= {\sigma_j}}^{\sigma_{j+1} -1} (\ell_{t} - \ell_{i_{\sigma_j}})
    \leq
    \frac 1\eta
      \sum_{j=0}^k \log \frac{\Pi_{M_{\sigma_{j+1}-1}}}{\pi_{i_{\sigma_j}}}
\end{equation}

\vspace{-2mm}
\noindent
since $\sigma_j = {\tau_{i_{\sigma_j}}}$, and where we denote $\Pi_M = \sum_{i=1}^M \pi_i$ for each $M \geq 1$. Taking $\pi_i = 1$, this bound reduces to $\frac 1\eta \sum_{j=0}^{k} \log M_{\sigma_{j+1}-1} \leq \frac 1\eta (k+1) \log M_T$. 
Taking $\pi_{i} = 1/{m_{\tau_i}}$, %%%%
%%%% $\pi_{i} = \frac{1}{\tau_i m_{\tau_i}}$ 
%%%% and further bounding $\Pi_{M_{\sigma_{j+1}-1}} \leq \Pi_{M_T} \leq  \sum_{t=1}^{T} \frac 1t \leq 1+ \log T$, 
so that $\Pi_{M_t} = t$, and further bounding $\Pi_{M_{\sigma_{j+1}-1}} = {\sigma_{j+1}-1} \leq \sigma_{j+1}$ for $0 \leq j \leq k-1$ and $\Pi_{M_{\sigma_{k+1}-1}} = T$, %%%%
we recover the bound~(\ref{eq:reg-fseq}) stated in the overview.

\vspace{-2mm}
\section{Growing experts and sequences of experts: the ``muting trick''}
\label{sec:grow-seq}

%%% Specialist viewpoint not enough !
%%% We should probably compress this !
\vspace{-1mm}
%% In Section~\ref{sec:grow-spec}, the interpretation of growing experts in the specialist setting enabled to control the regret with respect to sequences of fresh experts. This was achieved through the ``abstention trick'', which amounts to attributing to experts that have not entered yet predictions that do not affect the voting outcome.
Algorithm~\ghedge, based on the  specialist viewpoint, guarantees good regret bounds against \emph{fresh} sequences of experts and admits an efficient implementation. Instead of comparing only against fresh sequences of experts, it may be preferable to target \emph{arbitrary} admissible sequences of experts, that contain transitions to incumbent experts; 
%% that were already present for some rounds.
this could be beneficial when some experts start predicting well after a few rounds. A natural approach consists in applying the abstention trick to algorithms for a fixed expert set that target arbitrary sequences of experts (such as Fixed Share, see Appendix~\ref{ap:prop2}).
As it turns out, such an approach would require to maintain weights for unentered experts (which may be in unknown, even infinite, number in an anytime setting): the fact that one could obtain an efficient algorithm such as \ghedge{} was specific to the  exponential weights algorithm, and does not extend to more sophisticated algorithms that perform weight sharing.

%The fact that it only maintains weights for present experts, however, is specific to the simple exponential weights algorithm, and does not hold for more sophisticated algorithms that target different notions of regret.
%
In this section, we adopt a ``dual'' point of view, which proves more flexible. Indeed, in the growing expert ensemble setting, there are two ways to cope with the fact that some experts' predictions are undefined at each step. 
The abstention trick amounts to attributing \emph{predictions} to the experts which have not entered yet, so that they do not affect the learner's forecast. 
Another option %, which we explore here,
is to design a prior on \emph{sequences} of experts so that the \emph{weight} of unentered experts is $0$, and hence their predictions are irrelevant\footnote{In this case, the learner's predictions do not depend on the way we complete the experts' predictions, so the algorithm may be defined even when experts with zero weight do not output predictions.}; we call this the ``muting trick''.

After reviewing the well-known setting of aggregation of sequences of experts for a fixed set of experts (Section~\ref{sub:seq}) and presenting the generic algorithm \mhedge\ with its regret bound, we adapt it to the growing experts setting by providing  \fmhedge\ (Section~\ref{sub:fmhedge})  and \gmhedge\ (Section~\ref{sub:gmhedge}), that compete respectively with fresh and arbitrary sequences. 

%% In order to gain more flexibility, and in particular to control the regret with respect to arbitrary sequences of experts, we introduce in this section another point of view, which sees growing experts as sequences of experts. Specifically, it leads to another, somewhat ``dual'', way to deal with experts which have not entered yet: instead of attributing to them predictions that conform with those of the rest, we give them a weight equal to zero by carefully chosing the prior on \emph{sequences} of experts, so that their prediction is immaterial. We call this the ``muting trick''.

\vspace{-2mm}
\subsection{Aggregating sequences of experts}
\label{sub:seq}

\vspace{-1mm}
%% As we have seen, the exponential weights algorithm compares well to the best constant expert. A more ambitious problem, known as \emph{tracking the best expert}, consists in controlling the regret with respect to the best \emph{sequence} of experts with a limited number of shifts.
The problem of controlling the regret with respect to sequences of experts, known as \emph{tracking the best expert}, was introduced by \citet{hw98}, who proposed the simple \emph{Fixed Share} algorithm with good regret guarantees. 
A key fact, first recognized by~\cite{vovk99}, is that Fixed Share, and in fact many other weight sharing algorithms~\citep{kdr08,kdr13}, can be interpreted as the exponential weights on sequences of experts under a suitable prior. We will state this result in the general form of Lemma~\ref{lem:seq1}, which implies the regret bound of Proposition~\ref{prop:mhedge}.

\vspace{-1mm}
\paragraph{Markov prior.}

If $i^T = (i_1, \dots, i_T)$ is a finite sequence of experts, its predictions up to time $T$ are derived from those of the base experts $i \in \{ 1, \dots, M \}$ in the following way: $x_{t} (i^T) = x_{i_t,t} $ for $1\leq t \leq T$. Given a prior distribution $\pi = (\pi (i^T))_{i^T }$, we could in principle consider the exponentially weighted aggregation of sequences under this prior; however, such an algorithm would be intractable even for moderately low values of $T$, since it would require to store and update $O(M^T)$ weights.
Fortunately, when $\pi (i_1, \dots, i_T) = \theta_1 (i_1) \, \theta_2 (i_2 \cond i_1) \cdots \,\theta_T (i_T \cond i_{T-1})$ is a Markov probability distribution with initial measure $\bm\theta_1$ and transition matrices $\bm \theta_t$, $2 \leq t \leq T$, the exponentially weighted aggregation under the prior $\pi$ collapses to the efficient algorithm \mhedge.

\vspace{-1mm}
\begin{algorithm}
   \caption{\mhedge{} --- Aggregation of sequences of experts under a Markov prior}
\label{alg:seqm}              
\begin{algorithmic}[1]
  \STATE \textbf{Parameters:} Learning rate $\eta >0$, initial weights $\bm \theta_1 = (\theta_1(i))_{1 \leq i \leq M}$, and transition probabilities $\bm \theta_t = \big( \theta_{t} (i \cond j) \big)_{1\leq i,j \leq M}$ for all $t\geq 2$.
  \STATE \textbf{Initialization:} Set $\bm v_1 = \bm \theta_1$.
  \FOR{$t = 1, 2, \dots$} 
  \STATE Receive predictions $\bm x_t\in \X^M$ from the experts, and predict $x_t = \bm v_t \cdot \bm x_t$. 
  \STATE Observe $y_t \in \Y$, then derive the losses $\ell_t = \ell (x_t, y_t)$ and $\ell_{i,t} = \ell (x_{i,t}, y_t)$ and the posteriors

\vspace{-5mm}
  \begin{equation}
    \label{eq:mhedge-post}
    v_{i,t}^m 
    = \frac{v_{i,t}\, e^{-\eta\, \ell_{i,t}}}    
    {\sum_{j=1}^M v_{j,t}\, e^{-\eta\, \ell_{j,t}}} \, .
  \end{equation}
  
    \vspace{-1mm}
  \STATE Update the weights by   $\bm v_{t+1} = \bm \theta_{t+1} \, \bm v_{t}^m$, \ie
  
    \vspace{-6mm}
  \begin{equation}
    \label{eq:mhedge-share}
    v_{i,t+1} =
    \sum_{j=1}^M \theta_{t+1} (i \cond j) \, v_{j,t}^m \, .
  \end{equation}
  
   \vspace{-5mm}
  \ENDFOR
\end{algorithmic}
\end{algorithm}

\vspace{-3mm}
\begin{rem}
  Algorithm \mhedge\ only requires to store and update $O(M)$ weights. Due to the matrix product~(\ref{eq:mhedge-share}), the update may take a $O(M^2)$ time; however, all the transition matrices we consider lead to a simple update in $O(M)$ time.
\end{rem}

%Such ``bayesian'' %
%interpretations give more flexibility by allowing to tweak the prior, and  directly provide regret bounds through Proposition \ref{prop:exp}. However, for the classes of combinatorial experts we consider, the number of structured experts we aggregate typically  grows exponentially with the considered time horizon $T$. Therefore, in order to be able to implement such an algorithm, one has to show that it collapses to a fast 
%algorithm, provided we choose a suitable prior. 
%\paragraph{Markov prior: generic algorithm}
%% Since there are $M^T$ sequences of experts of length $T$, the naive implementation of the aggregation of sequences of experts has the prohibitive exponential complexity $O(M^T)$. In order to derive a practical algorithm, it remains to choose an appropriate prior $\mu$ so that the algorithm collapses to a more efficient one; as is turns out, this is achieved by Markov distributions $\mu$ on the sequences of experts, which only require to store and update $M$ weights~\citep{vovk99,kdr08,kdr13}: 
%%%: $(i)$ the algorithm simplifies to a more efficient one, and $(ii)$ the regret bound~\eqref{eq:reg-seq1} is small for regular sequences of experts.
%, \ie the weights concentrate on sequences which do not switch too often.

\begin{lemma}
  \label{lem:seq1}
  For every $T \geq 1$, the forecasts of algorithm \mhedge\ coincide up to time $T$ with those of the exponential aggregation of {finite} sequences of experts $i^T = (i_1, \dots , i_T)$ under the Markov prior with initial distribution $\bm\theta_1$ and transition matrices $\bm \theta_2, \dots, \bm\theta_T$.
\end{lemma}
\vspace{-2mm}
\noindent
Lemma~\ref{lem:seq1} -- proven in Appendix~\ref{ap:prop2} -- and Proposition~\ref{prop:exp} directly imply the following regret bound.

\begin{proposition}
  \label{prop:mhedge}
  Algorithm~\aggmarkov, with initial distribution $\bm \theta_1$ and transition matrices $\bm \theta_t$,  guarantees the following regret bound: for every $T \geq 1$ and any sequence of experts $%i^T =
  (i_1, \dots, i_T)$,
  \begin{equation}
    \label{eq:reg-mhedge}
    \sum_{t=1}^T \ell_t - \sum_{t=1}^T \ell_{i_t,t}
    \leq \frac 1\eta \log \frac 1{\theta_1({i_1})}
    + \frac 1\eta \sum_{t=2}^T \log \frac{1}{\theta_t (i_t \cond i_{t-1})} \, .  
  \end{equation}
\end{proposition}
It is worth noting that the transition probabilities $\bm \theta_t$ only intervene at step $t$ in algorithm \mhedge, and hence they can be chosen at this time.
\vspace{-1mm}
\paragraph{Notable examples.}
In Appendix~\ref{ap:prop2}, we discuss particular instances of \aggmarkov{} that lead to well-known algorithms %from the literature
(such as Fixed Share), and recover their regret bounds using Proposition~\ref{prop:mhedge}.

%%% say somewhere that \fmhedge\ is a particular case of the general algorithm \gmhedge
\vspace{-2mm} 
\subsection{Application to sequences of fresh experts}
\label{sub:fmhedge}

We now explain how to specify the generic algorithm \mhedge\ in order to adapt it to the growing experts setting. This adaptation relies on the ``muting trick'': to obtain a strategy which is well-defined for growing experts, one has to ensure that experts who do not predict have zero weight, which amounts to saying that all weight is put to \emph{admissible} sequences of experts.
Importantly, this is possible even when the numbers $M_t$ are not known from the beginning, since \emph{the transition matrices $\bm \theta_t$ can be chosen at time $t$}, when $M_t$ is revealed.

We start in this section by designing an algorithm \fmhedge\ that compares to sequences of fresh experts; 
to achieve this, it is natural to design a prior that assigns full probability to sequences of fresh experts.
It turns out that we can recover an algorithm similar to the algorithm~\ghedge, with the same regret guarantees, through this seemingly different viewpoint.

Let $\bm \pi = (\pi_i)_{i \geq 1}$ be an \emph{unnormalized prior} as in Section~\ref{sub:ghedge}. For each $M \geq 1$, we denote $\Pi_M = \sum_{i=1}^M \pi_i$.
%% ; in particular, when $\pi_i = \frac{\nu_{\tau_i}}{m_{\tau_i}}$ we have $\Pw_{M_t} = \Nu_t = \sum_{s=1}^t \nu_s$.
We consider the following transition matrices $\bm \theta_t$ in strategy \mhedge:
\begin{equation}
  \label{eq:trans-fmhedge}
  \theta_1 (i) = \frac{\pi_i}{\Pi_{M_1}} \bm 1_{i \leq M_1}
  \quad ; \quad
  \theta_{t+1} (i \cond j) = \frac{\Pi_{M_t}}{\Pi_{M_{t+1}}} \bm 1_{i=j}
  + \frac{\pi_i}{\Pi_{M_{t+1}}} \bm 1_{M_t +1 \leq i \leq M_{t+1}}
\end{equation}
for every $i \geq 1$, $t \geq 1$ and $j \in \{ 1, \dots, M_t \}$. The other transition probabilities $\theta_{t+1} (i \cond j)$ for $j >M_t$ are irrelevant; indeed, a simple induction shows that $v_{j,t} = 0$ for every $j > M_t$, so that the instantiation of algorithm \mhedge\ with the transition probabilities~(\ref{eq:trans-fmhedge}) leads to the forecasts
\begin{equation}
  \label{eq:pred-fmhedge}
  x_t = \sum_{i=1}^{M_t} v_{i,t} \, x_{i,t}
\end{equation}
(which do not depend on the undefined prediction of the experts $i >M_t$)
where the weights $(v_{i,t})_{1\leq i \leq M_t}$ are recursively defined by $v_{i,1} = \frac{\pi_i}{\Pi_{M_1}}$ $(1 \leq i \leq M_1)$ and the update
\begin{equation}
  \label{eq:upd-fmhedge}
  v_{i,t+1} = \frac{\Pi_{M_t}}{\Pi_{M_{t+1}}} \,  v_{i,t}^m
  \quad (1 \leq i \leq M_t) \, ;
  \quad
  v_{i,t+1} = \frac{\pi_i}{\Pi_{M_{t+1}}}
  \quad (M_t +1 \leq i \leq M_{t+1}) \, ,
\end{equation}
where we set $v_{i,t}^m = \displaystyle \frac{v_{i,t}\, e^{-\eta\, \ell_{i,t}}}{\sum_{j=1}^{M_t} v_{j,t} \,  e^{-\eta\, \ell_{j,t}}}$ for $1 \leq i \leq M_t$. We call this algorithm \fmhedge.
%%% Note that, like the \ghedge\ algorithm, \fmhedge\ has a complexity of $O (M_t)$ at round $t$, which grows linearly with the number of experts.
%
\vspace{-3mm}
\begin{theorem}
  \label{thm:fmhedge}
  Algorithm \fmhedge\ using weights $\bm \pi$ achieves the following regret bound: for every $T\! \geq\! 1$ and sequence of fresh experts $i^T\!  =\! (i_1, \dots, i_T)$ with shifts at times $\bm \sigma\! =\! (\sigma_1 , \dots, \sigma_k)$,
  
  \vspace{-5mm}
  \begin{equation}
    \label{eq:reg-fmhedge}
    L_T - L_T (i^T)
    \leq
    \frac 1\eta \sum_{j=0}^k \log \frac{1}{\pi_{i_{\sigma_j}}}
    + \frac 1\eta \sum_{j=1}^k \log \Pi_{M_{\sigma_{j}-1}}
    + \frac 1\eta \log \Pw_{M_T}
    \, .
  \end{equation}  
  Additionally, the time and space complexity of the algorithm at each time step $t \geq 1$ is $O(M_t)$.
\end{theorem}

\vspace{-2mm}
\begin{myproof}
  For any sequence of fresh experts $i^T \in \fseq_T (\bm \sigma)$, replacing in the bound~(\ref{eq:reg-mhedge}) of Proposition~\ref{prop:mhedge} the conditional probabilities $\theta_{t+1} (i_{t+1} \cond i_t)$ by their values (defined by~(\ref{eq:trans-fmhedge})), we get
%  This is a direct consequence of Proposition~\ref{prop:mhedge}, which gives for any sequence of fresh experts $i^T \in \fseq_T (\bm \sigma)$, replacing in the bound~(\ref{eq:reg-mhedge}) the conditional probabilities $\theta_{t+1} (i_{t+1} \cond i_t)$ by their values (see~(\ref{eq:trans-fmhedge})):
  \begin{equation*}
    L_T - L_T (i^T)
  \leq \frac 1\eta \sum_{j=0}^k
  \left\{ \log \bigg( \frac 1{\pi_{i_{\sigma_j}}} \Pi_{M_{\sigma_j}} \bigg)
    + \sum_{t=\sigma_j+1}^{\sigma_{j+1}-1} \log \frac{\Pi_{M_t}}{\Pi_{M_{t-1}}}
  \right\} \\
  = \frac 1\eta \sum_{j=0}^k \log 
    \frac {\Pi_{M_{\sigma_{j+1}-1}}}{\pi_{i_{\sigma_j}}} 
  \end{equation*}
  which is precisely the desired bound~(\ref{eq:reg-fmhedge}).
\end{myproof}
\vspace{-3mm}
\begin{rem}
  The regret bound~(\ref{eq:reg-fmhedge}) of the \fmhedge\ algorithm against sequences of fresh experts is exactly the same as the one of the \ghedge\ algorithm~(\ref{eq:reg-fresh-ghedge}). This is not a coincidence: the two algorithms are almost identical, except that expert $i$ is introduced with a weight $\pi_i/(\sum_{i=1}^{M_{\tau_i}} \pi_i)$ by \fmhedge\ and $\pi_i e^{-\eta\, L_{\tau_i-1}}/(\sum_{j=1}^{M_{\tau_i}} \pi_j e^{-\eta\, L_{j,\tau_i-1}} )$ by \ghedge. In the case of the logarithmic loss (with $\eta =1$), these two weights are equal (see Remark~\ref{rem:equality}), and hence the strategies \ghedge\ and \fmhedge\ coincide.
%
%%  Indeed, we saw that \ghedge\ coincides up to time $\tau_i$ with the specialist aggregation of $j=1, \dots, \tau_i$ under the prior $\pi_j/\Pi_{M_{\tau_i}}$
%  Indeed, since in the case of the logarithmic loss $x \mapsto e^{-\ell (x,y)} = x(y)$ is linear, algorithm \ghedge\ satisfies for every $t= 1, \dots, \tau_i$: $e^{-\ell_t} = (\sum_{i=1}^{M_t} w_{i,t} e^{-\ell_{i,t}}) / (\sum_{i=1}^{M_t} w_{i,t})= (\sum_{i=1}^{M_{\tau_i}} w_{i,t} e^{-\ell_{i,t}}) / (\sum_{i=1}^{M_{\tau_i}} w_{i,t}) = W_{t+1}/W_t$ where $W_t := \sum_{i=1}$, and hence by taking the product over $t=1, \dots,\tau_i-1$.
%  
% in the case of the logarithmic loss, this is the same, see remark in the specialists section: the weights of inactive specialists are left unchanged
\end{rem}

%% In particular, we recover the bound of the overview.
%% In the case when $\pw_i = \frac{\nu_{\tau_i}}{m_{\tau_i}}$, $\Pi_{M_{t}} = \Nu_t = \sum_{s=1}^t \nu_s $. In particular, $\nu_t = 1$, $\nu_t = \frac 1t$.

\vspace{-3mm}
\subsection{Regret against arbitrary sequences of experts}
\label{sub:gmhedge}

\vspace{-1mm}
We now consider the more ambitious objective of comparing to \emph{arbitrary} admissible sequences of experts. This can be done by using another choice of transition matrices, which puts all the weight to admissible sequences of experts (and not just sequences of fresh experts).

Algorithm~\gmhedge\ %%, which is a generalisation of \fmhedge,
instantiates \mhedge\ on the transition matrices
\begin{equation}
  \label{eq:trans-gmhedge}
  \theta_1 (i) = \frac{\pi_i}{\Pi_{M_1}} \bm 1_{i \leq M_1}
  \quad ; \quad
  \theta_{t+1} (i \cond j) = \alpha_{t+1} \frac{\pi_i}{\Pi_{M_{t+1}}} + (1-\alpha_{t+1}) \, \theta_{t+1}^{(f)} (i \cond j)
  %% =  \alpha_{t+1} \frac{\pi_i}{\Pi_{M_{t+1}}}
  %% + (1- \alpha_{t+1}) \frac{\Pi_{M_t}}{\Pi_{M_{t+1}}} \bm 1_{i=j}
%%  + (1- \alpha_{t+1})\frac{\pi_i}{\Pi_{M_{t+1}}} \bm 1_{M_t +1 \leq i \leq M_{t+1}}
\end{equation}

\vspace{-1mm}
\noindent
where $\bm \theta_{t}^{(f)}$ denote the transition matrices of algorithm~\fmhedge.
As before, this leads to a well-defined growing experts algorithm which predicts 
$  x_t = \sum_{i=1}^{M_t} v_{i,t} \, x_{i,t}$,
where the weights $(v_{i,t})_{1\leq i \leq M_t}$ are recursively defined by $v_{i,1} = \frac{\pi_i}{\Pi_{M_1}}$ $(1 \leq i \leq M_1)$ and the update
\begin{equation}
  \label{eq:upd-gmhedge}
  v_{i,t+1} = (1-\alpha_{t+1})\frac{\Pi_{M_t}}{\Pi_{M_{t+1}}} \,  v_{i,t}^m
  + \alpha_{t+1}\frac{\pi_i}{\Pi_{M_{t+1}}}
%  \ 
%  \mbox{ with } \
%  v_{i,t}^m
%  =
%%  \frac{\Pi_{M_t}}{\Pi_{M_{t+1}}} %
%%  \frac{v_{i,t}\, e^{-\eta\, \ell_{i,t}}}{\sum_{j=1}^{M_t} v_{j,t} \,  e^{-\eta\, \ell_{j,t}}}
  \ (1 \leq i \leq M_t) \, ;
  \
  v_{i,t+1} = \frac{\pi_i}{\Pi_{M_{t+1}}}
  \ (M_t +1 \leq i \leq M_{t+1}) \, ,
\end{equation}

\vspace{-3mm}
\noindent
where again $v_{i,t}^m = \displaystyle \frac{v_{i,t}\, e^{-\eta\, \ell_{i,t}}}{\sum_{j=1}^{M_t} v_{j,t} \,  e^{-\eta\, \ell_{j,t}}}$ for $1 \leq i \leq M_t$. In this case, Proposition~\ref{prop:mhedge} yields:
\begin{theorem}
  \label{thm:gmhedge}
  Algorithm \gmhedge\ based on the weights $\bm \pi$ and parameters $(\alpha_{t})_{t\geq 2}$ achieves the following regret bound: for every $T \geq 1$, and every admissible sequence of experts $i^T  = (i_1, \dots, i_T)$ with shifts at times $\bm \sigma = (\sigma_1 , \dots, \sigma_k)$,
  \begin{equation}
    \label{eq:reg-gmhedge}
      L_T - L_T (i^T)
  \leq
  \frac 1\eta \left\{
    \sum_{j=0}^{k} \log \frac{\Pi_{M_{\sigma_{j+1}-1}}}{\pi_{i_{\sigma_j}}}
    + \sum_{j=1}^{k_1} \log \frac 1{\alpha_{\sigma^1_j}}
    + \sum_{2\leq t \leq T \pp t \not\in \bm \sigma} \log \frac{1}{1-\alpha_t}
  \right\}
%
%    L_T - \inf_{i^T \in \aseq_T (\bm \sigma^0 ; \bm \sigma^1)} L_T (i^T)
%    \leq
%    \frac 1\eta \sum_{j=0}^k \log \frac{1}{\pi_{i_{\sigma_j}}}
%    + \frac 1\eta \sum_{j=1}^k \log \Pi_{M_{\sigma_{j}-1}}
%    + \frac 1\eta \log \Pw_{M_T}
    \, .
  \end{equation}
  where  $\bm \sigma^0 = (\sigma^0_1 , \dots, \sigma^0_{k_0})$ (resp. $\bm \sigma^1 = (\sigma^1_1 , \dots, \sigma^1_{k_1})$) denotes the shifts to fresh  (resp. incumbent) experts, with $k=k_0+k_1$.
  Moreover, it has a $O(M_t)$ time and space complexity at each step $t \geq 1$.
\end{theorem}

\vspace{-3mm}
 \begin{rem}
Note that by choosing $\alpha_t = \frac 1t$, we have, since $\frac 1{1- 1/t} = \frac t{t-1}$,
  \begin{equation*}
  \sum_{j=1}^{k_1} \log \frac 1{\alpha_{\sigma^1_j}}
  + \sum_{2\leq t \leq T \pp t \not\in \bm \sigma} \log \frac{1}{1-\alpha_t}
  \leq   \sum_{j=1}^{k_1} \log \sigma^1_j + \sum_{t=2}^T \log \frac{t}{t-1}
  =  \sum_{j=1}^{k_1} \log \sigma^1_j + \log T \, .
  \end{equation*}
  Additionally, by setting $\pi_i = 1$ the bound~(\ref{eq:reg-gmhedge}) becomes
  $  \frac 1\eta (
  \sum_{j=0}^{k} \log M_{\sigma_{j+1}-1}
  + \sum_{j=1}^{k_1} \log \sigma^1_j + \log T 
  )$, which is lower than $\frac 1\eta (k+1) \log M_T + \frac 1\eta (k_1+1) \log T$. We can also recover the bound~(\ref{eq:reg-aseq}) by setting $\pi_i = \frac 1{\tau_i m_{\tau_i}}$, since in this case we have $\Pi_{M_{\sigma_{j+1}-1}} \leq \Pi_{M_T} \leq \sum_{t=1}^T \frac 1t \leq 1+\log T$.
  %% (This last derivation was already done before: simply refer to it ?)  
\end{rem}

\vspace{-6mm}
\section{Combining growing experts and sequences of sleeping experts}
\label{sec:grow-sleep}

\vspace{-1mm}
Sections~\ref{sec:grow-spec} and~\ref{sec:grow-seq} studied the problem of growing experts using tools from two different settings (specialists and sequences of experts). Drawing on ideas from~\cite{kool}, we show in this section how to combine these two frameworks, in order to address the more challenging problem of controlling the regret with respect to \emph{sparse sequences of experts}  in the growing experts setting. Note that the refinement to sparse sequences of experts is particularly relevant in the context of a growing experts ensemble, since in this context the total number of experts will typically be large.
%
% this is typically the case in the growing experts setting, where $M_T$ is typically at least as large as $T$.

\vspace{-3mm}
\subsection{Sleeping experts: generic result}
\label{sub:sleep}

\vspace{-1mm}
The problem of comparing to sparse sequences of experts, or \emph{tracking a small pool of experts}, is a refinement on the problem of tracking the best expert.
The seminal paper~\citep{mpp} proposed an ad-hoc strategy with essentially optimal regret bounds, %; (given a bound on the number of shifts of the comparison sequences);
the \emph{Mixing Past Posteriors} (MPP) algorithm
(see  also \cite{gaill}). A full ``bayesian'' interpretation of this algorithm in terms of the aggregation of ``sleeping experts'' was given by~\cite{kool}, which enabled the authors to propose a more efficient alternative. Here, by  reinterpreting this construction, we propose a % slightly
more general algorithm and regret bound (Proposition~\ref{prop:reg-sleep}); this extension will be crucial to adapt this strategy to the growing experts setting (Section~\ref{sub:grow-sparse}).

% Derive it fast: aggregation of sleeping experts under a Markov prior

Given a fixed set of {experts} $\{ 1, \dots, M\}$, we call \emph{sleeping expert} a couple $ %\iota =
(i,a) \in \{ 1, \dots, M \} \times \{ 0,1 \}$; we endow the set of sleeping experts  with a {specialist} structure by deciding that $(i, a)$ is active if and only if $a=1$, and that $x_t (i,1) := x_{i,t}$ is the prediction of expert $i$.
A key insight from~\cite{kool} is to decompose the regret with respect to a sparse sequence $i^T = (i_1, \dots, i_T)$ of experts, taking values in the set $\{ e_p \mid 1\leq p \leq n \}$, in the following way:
\begin{equation*}
  \sum_{t=1}^T (\ell_t - \ell_{i_t,t})
  = \sum_{p=1}^n \sum_{t \leq T \pp i_t=e_p} (\ell_t - \ell_{e_p,t})
  = \sum_{p=1}^n \sum_{t=1}^T (\ell_t - \ell_t (e_p, a_{p,t}))
  = n \sum_{\iota^T} u (\iota^T) (L_T - L_T (\iota^T))
\end{equation*}
where $a_{p,t} := \bm 1_{i_t = e_p}$, and $u$ is the probability distribution on the sequences $\iota^T$ of sleeping experts which is uniform on the $n$ sequences $\iota_p^T = (e_p , a_{p,t})_{1\leq t \leq T}$, $p=1, \dots, n$.
Note that in the second equality we used the ``abstention trick'', which  attributes to inactive sleeping experts $(e_p,0)$ the prediction $x_t$ of the algorithm.

We can now aggregate sequences of sleeping experts under a Markov prior, given initial weights $\theta_1 (i,a)$ and transition probabilities $\theta_{t+1} ( i_{t+1}, a_{t+1} \cond i_t,a_t )$, recalling that $\bm\theta_t$ can be chosen at step $t$. For convenience, we restrict here to transitions that only occur between sleeping experts $(i,a)$ with the same base expert, and denote $\theta_{i,t} (a \cond b) = \theta_t (i,a \cond i,b)$ for $a,b \in \{ 0,1 \}$. This leads to the algorithm \smhedge.
\begin{rem}
  The structure of our prior is slightly more general than the one used by~\cite{kool}, which considered priors on couples $(i, a^T)$ with an independence structure: $\pi (i, a^T) = \pi (i) \, \pi (a^T)$, with $\pi (a^T)$ a Markov distribution, which amounts to saying that the transition probabilities $\theta_{i,t} (a \cond b)$ could not depend on $i$. This additional flexibility will enable in Section~\ref{sub:grow-sparse}  the ``muting trick'', which allows to convert \smhedge\ to the growing experts setting.
  
Additionally, allowing transitions between sleeping experts $(i,1)$ and $(j,1)$ for $i \neq j$ may be interesting in its own right, e.g. if one seeks to control \emph{at the same time} the regret with respect to sparse and non-sparse sequences of experts.
%%%%  The regret with respect to sparse sequences may be controlled through the decomposition indicated above (\ie along sequences of the form $(i,a_t)_{t\geq 1}$), whereas the regret with respect to non-sparse sequences may be controlled in the same way as in Section~\ref{sub:seq} (along sequences of the form $(i_t,1)_{t\geq 1}$). 
\end{rem}

  \begin{algorithm}            
  \caption{\smhedge: sequences of sleeping experts under a Markov chain prior}
\label{alg:sleep}              
\begin{algorithmic}[1]
  \STATE \textbf{Parameters:} Learning rate $\eta >0$, (normalized) prior $\bm \pi$ on the experts, initial wake/sleep probabilities $\theta_{i,1} (a)$, transition probabilities $\bm \theta_{i,t} = \big( \theta_{i,t} (a \cond b ) \big)_{a,b \in \{0,1\}}$ for $t\geq 2, \, 1\leq i \leq  M$.
  \STATE \textbf{Initialization:} Set $v_1 (i,a) = \pi_i \, \theta_{i,1} (a)$ for $i= 1, \dots, M$ and $a \in \{0,1\}$.
  \FOR{$t = 1, 2, \dots$} 
  \STATE Receive predictions $\bm x_t\in \X^M$ from the experts, and predict
  \begin{equation}
    \label{eq:pred-sleep}
    x_t = \frac{\sum_{i=1}^M v_t(i,1) \, x_{i,t}}{\sum_{i=1}^M v_t(i,1)} \, .
  \end{equation}
    \vspace{-2mm}
  \STATE Observe $y_t \in \Y$, then derive the losses $\ell_t (i,0) = \ell_t = \ell (x_t, y_t)$, $\ell_t (i,1) = \ell_{i,t} = \ell (x_{i,t}, y_t)$ and the posteriors 
  \vspace{-4mm}
  \begin{equation}
    \label{eq:post-sleep}
    v_{t}^m (i,a)
    = \frac{v_t (i,a) \, e^{-\eta\, \ell_t (i,a)}}    
    {\sum_{i', a'} v_t (i',a') \, e^{-\eta\, \ell_t (i',a')}} \, .
  \end{equation}
  \vspace{-2mm}
  \STATE Update the weights by
  \vspace{-4mm}
  \begin{equation}
    \label{eq:sl-update2}
 v_{t+1} (i,a) = \sum_{b\in \{0,1\}} \theta_{i,t+1} (a \cond b) \, v_{t}^m (i,b) \, .
  \end{equation}

  \vspace{-3mm}
  \ENDFOR
\end{algorithmic}
\end{algorithm}

\vspace{-4mm}
\begin{proposition}
  \label{prop:reg-sleep}
  Strategy~\smhedge\ guarantees the following regret bound: for each sequence $i^T$ of experts taking values in the pool $\{ e_p \mid 1 \leq p \leq n \}$, denoting $a_{p,t} = \bm 1_{i_t = e_p}$
  \begin{equation}
    \label{eq:reg-sleep}
    L_T  - L_T (i^T)
  \leq \frac 1\eta \sum_{p=1}^n \left( \ln \frac{1/n}{\pi_{e_p}}
    + \ln \frac 1{\theta_{e_p,1} (a_{p,1})} + \sum_{t=2}^T \ln \frac 1 { \theta_{e_p,t} (a_{p,t} \cond a_{p,t-1})}
  \right)
  \, .    
  \end{equation}
\end{proposition}
The proof of Proposition~\ref{prop:reg-sleep} is given in Appendix~\ref{ap:sleep}.

\vspace{-3mm}

\subsection{Sparse shifting regret for growing experts}
%%\subsection{Sparse sequences of experts}
\label{sub:grow-sparse}

We show here how to instantiate algorithm \smhedge\ in order to adapt it to the growing experts setting. Again, we use a ``muting trick'' which attributes a zero weight to experts that have not entered.

Let us consider prior weights $\bm \pi = (\pi_i)_{i \geq 1}$ on the experts, which may be unnormalized and chosen at entry time.
Let $\alpha_{t},\beta_{t} \in (0,1)$ for $t\geq 2$.
We set $\theta_{i,1} (1) = \frac 12$ for $i = 1, \dots, M_1$ and $0$ otherwise; moreover, for every $t\geq 1$, we take $\theta_{i,t+1} (1 \cond \cdot) = 0$ for $i>M_{t+1}$ (recall that $\bm\theta_{i,t+1}$ can be chosen at step $t+1$), $\theta_{i,t+1} (1 \cond \cdot) = \frac 12$ if $M_{t} +1 \leq i \leq M_{t+1}$, and for $i \leq M_t$: $\theta_{i,t+1} (0 \cond 1) = \alpha_{t+1}$, $\theta_{i,t+1} (1 \cond 0) = \beta_{t+1}$.
The algorithm obtained with these choices, which we call \gsmhedge, is well-defined and predicts
$x_t = %\frac
({\sum_{i=1}^{M_t} v_t(i,1) \, x_{i,t}})/({\sum_{i=1}^{M_t} v_t(i,1)})$,
where the weights $(v_t(i,a))_{1\leq i \leq M_t, \, a \in \{ 0,1\}}$ are defined by $v_1 (i,a) = \frac 12 \, \pi_i$ $(1\leq i \leq M_1)$ and by the update
\begin{equation*}
  v_{t+1} (i,a) = \sum_{b\in \{0,1\}} \theta_{i,t+1} (a \cond b) \, v_{t}^m (i,b)
  \ (1 \leq i \leq M_t %, a\in \{ 0, 1\}
  ) \, ;
  \
  v_{t+1} (i,a) = \frac 12\, \pi_i
  \ (M_t +1 \leq i \leq M_{t+1}) \, ,
\end{equation*}
with
 $   v_{t}^m (i,a)
     = {v_t (i,a) \, e^{-\eta\, \ell_t (i,a)}}    /
     {\sum_{i=1}^{M_t}\sum_{a'\in \{0,1\}} v_t (i',a') \, e^{-\eta\, \ell_t (i',a')}} 
     $ for $1 \leq i \leq M_t$.
% $v_1 (i,a) = \frac 12 \, \pi_i\,  \bm 1_{i \leq M_1}$ and the update at time $t+1$: for $1\leq i \leq M_{t}$, 
% $ v_{t+1} (i,a) = \sum_{b\in \{0,1\}} \theta_{i,t+1} (a \cond b) \, v_{t}^m (i,b)$  where
% $  \displaystyle  v_{t}^m (i,a)
%     = \frac{v_t (i,a) \, e^{-\eta\, \ell_t (i,a)}}    
%     {\sum_{i=1}^{M_t}\sum_{a'\in \{0,1\}} v_t (i',a') \, e^{-\eta\, \ell_t (i',a')}} 
%     $; and for $M_t +1 \leq i \leq M_{t+1}$, $v_{t+1} (i,a) = \frac 12\, \pi_i$.
%%%    $\bullet$ The initialization $v_1 (i,a) = \frac 12 \, \pi_i$ for $1\leq i \leq M_1$.
%%%    $\bullet$ The update at step $t+1$: for $1\leq i \leq M_{t}$, $ v_{t+1} (i,a) = \sum_{b\in \{0,1\}} \theta_{i,t+1} (a \cond b) \, v_{t}^m (i,b)$  where
%%%$  \displaystyle  v_{t}^m (i,a)
%%%    = \frac{v_t (i,a) \, e^{-\eta\, \ell_t (i,a)}}    
%%%    {\sum_{i=1}^{M_t}\sum_{a'\in \{0,1\}} v_t (i',a') \, e^{-\eta\, \ell_t (i',a')}} 
%%%    $; and for $M_t +1 \leq i \leq M_{t+1}$, $v_{t+1} (i,a) = \frac 12 \pi_i$.

\begin{theorem}
  \label{thm:gsmhedge}
  Algorithm~\gsmhedge\ guarantees the following% regret bound
  : for each $T \geq 1$ and any sequence $i^T$ of experts taking values in the pool $\{ e_p \mid 1 \leq p \leq n \}$, denoting $a_{p,t} = \bm 1_{i_t = e_p}$
  \begin{align}
    \label{eq:reg-gsmhedge}
    L_T  - L_T (i^T)
       &\leq
      \frac 1\eta \sum_{p=1}^n \ln \frac{\Pi_{M_T}/n}{\pi_{e_p}}
        + \frac 1\eta n \log 2 +        
        \frac 1\eta \sum_{t=2}^T \left[ \log \frac 1{1-\alpha_t} + (n-1) \log \frac 1{1-\beta_t} \right] \nonumber \\
       &+ \frac 1\eta \sum_{j=1}^k \left( \log \frac{1}{\alpha_{\sigma_j}} + \ln
          \frac 1{\beta_{\sigma_j}} \right)
  \end{align}
  where $\bm \sigma = \sigma_1 < \dots < \sigma_k$ denote the shifting times of $i^T$.
  Moreover, the algorithm has a $O(M_t)$ time and space complexity at step $t$, for every $t\geq 1$.
\end{theorem}

\vspace{-1mm}
 In particular, Theorem~\ref{thm:gsmhedge} enables to recover the bound~\eqref{eq:reg-last} for $\alpha_t = \beta_t = \frac 1t$ and $\pi_i = \frac{1}{\tau_i m_{\tau_i}}$.

 \vspace{1mm}
%% mettre la preuve en annexe, et la rédiger avec un peu plus en détail ?
\begin{myproof}
  Note that algorithm~\gsmhedge\ is invariant under any change of prior $\bm \pi \leftarrow \lambda \bm \pi$ due to the renormalisation in the formula defining $x_t$. In particular, setting $\lambda = 1/\Pi_{M_T}$, we see that it coincides up to time $T$ with algorithm \smhedge\ with set of experts
  $\{ 1, \dots , M_T \}$ and (normalized) prior weights $\pi_i / \Pi_{M_T}$. The bound~(\ref{eq:reg-gsmhedge})
  is now a consequence of the general regret bound~(\ref{eq:reg-sleep}), by substituting for the values of $\bm \theta_{i,t+1}$.
\end{myproof}

\medskip
\noindent{\bf \large{Conclusion.\ }}
%% Improve it?
In this paper, we extended aggregation of experts to the
\textit{growing expert} setting, where novel experts are made available
at any time. In this context when the set of experts itself varies, it is natural to seek to track the best expert; different comparison classes of increasing complexity were considered. 
In order to obtain efficient algorithms with a per-round complexity linear in the current number of experts, we started with generic reformulation of existing algorithms for fixed expert set, and identified two orthogonal techniques (the ``abstention trick'' from the specialist literature, and the ``muting trick'') to adapt them to sequentially incoming forecasters. Combined with a proper tuning of the parameters of the prior, this enabled us to obtain tight regret bounds, adaptive to the parameters of the comparison class.
%
% We revisited standard algorithms and showed that a non
% trivial adaptation of them is required in order to obtain a controlled
% error.  Using the ``abstention trick'' from the setting of specialists
% and the somewhat dual ``muting trick'' that we introduce, we derive novel
% algorithms enjoying near optimal regret bounds for increasingly
% challenging notions of regret, while maintaining each time a small
% computational complexity, that grows linearly with the number of present experts. % (e.g. thanks to the use of Markov priors).
Along the way, we recovered several key results from the literature as
special case of our analysis, in a somewhat unified approach.

Although we considered the exp-concave assumption to avoid distracting
the reader from the main challenges of the growing expert setting,
extending our results to the bounded convex case in which the
parameter $\eta$ needs to be adaptively tuned seems possible and is left for future work.
In addition, building on the recent work of \cite{jun17a} might bring further improvements in this case.
Another natural extension of our work would be to address the same questions in the framework of online convex optimization~\citep{oloco,oco}, when the gradient of the loss function is made available at each time step.

% Acknowledgments---Will not appear in anonymized version
\acks{This work has been supported by the French Agence Nationale de la Recherche (ANR), under grant ANR-16- CE40-0002 (project BADASS), and Inria. JM acknowledges support from \'{E}cole Polytechnique fund raising -- Data Science Initiative.}

\bibliography{biblio-growing}

\begin{thebibliography}{33}
\providecommand{\natexlab}[1]{#1}
\providecommand{\url}[1]{\texttt{#1}}
\expandafter\ifx\csname urlstyle\endcsname\relax
  \providecommand{\doi}[1]{doi: #1}\else
  \providecommand{\doi}{doi: \begingroup \urlstyle{rm}\Url}\fi

\bibitem[Bousquet and Warmuth(2002)]{mpp}
Olivier Bousquet and Manfred~K. Warmuth.
\newblock Tracking a small set of experts by mixing past posteriors.
\newblock \emph{The Journal of Machine Learning Research}, 3:\penalty0
  363--396, 2002.

\bibitem[Cesa-Bianchi and Lugosi(2006)]{PLG}
Nicol{\`o} Cesa-Bianchi and G{\'a}bor Lugosi.
\newblock \emph{Prediction, Learning, and Games}.
\newblock Cambridge University Press, Cambridge, New York, USA, 2006.

\bibitem[Cesa-Bianchi et~al.(2012)Cesa-Bianchi, Gaillard, Lugosi, and
  Stoltz]{gaill}
Nicol{\`o} Cesa-Bianchi, Pierre Gaillard, G{\'a}bor Lugosi, and Gilles Stoltz.
\newblock Mirror descent meets fixed share (and feels no regret).
\newblock In \emph{Advances in Neural Information Processing Systems 25}, pages
  980--988. Curran Associates, Inc., 2012.

\bibitem[Chernov and Vovk(2009)]{cv}
Alexey Chernov and Vladimir Vovk.
\newblock Prediction with expert evaluators' advice.
\newblock In \emph{Proceedings of the 20th international conference on
  Algorithmic Learning Theory}, ALT~'09, pages 8--22, Berlin, Heidelberg, 2009.
  Springer-Verlag.

\bibitem[de~Rooij et~al.(2014)de~Rooij, van Erven, Gr{\"u}nwald, and
  Koolen]{ftl}
Steven de~Rooij, Tim van Erven, Peter Gr{\"u}nwald, and Wouter~M. Koolen.
\newblock Follow the leader if you can, hedge if you must.
\newblock \emph{Journal of Machine Learning Research}, 15:\penalty0 1281--1316,
  2014.

\bibitem[Freund et~al.(1997)Freund, Schapire, Singer, and Warmuth]{fss}
Yoav Freund, Robert~E. Schapire, Yoram Singer, and Manfred~K. Warmuth.
\newblock Using and combining predictors that specialize.
\newblock In \emph{Proceedings of the 29th Annual ACM Symposium on Theory of
  Computing (STOC)}, pages 334--343, 1997.

\bibitem[Gofer et~al.(2013)Gofer, Cesa-Bianchi, Gentile, and
  Mansour]{gofer2013branching}
Eyal Gofer, Nicol{\`o} Cesa-Bianchi, Claudio Gentile, and Yishay Mansour.
\newblock Regret minimization for branching experts.
\newblock In \emph{Proceedings of the 26th Annual Conference on Learning Theory
  (COLT)}, pages 618--638, 2013.

\bibitem[Gy{\"o}rfi et~al.(1999)Gy{\"o}rfi, Lugosi, and Morvai]{gyo99}
L{\'a}szl{\'o} Gy{\"o}rfi, G{\'a}bor Lugosi, and Gust{\'a}v Morvai.
\newblock A simple randomized algorithm for sequential prediction of ergodic
  time series.
\newblock \emph{IEEE Transactions on Information Theory}, 45\penalty0
  (7):\penalty0 2642--2650, 1999.

\bibitem[Gyorgy et~al.(2012)Gyorgy, Linder, and Lugosi]{gyorgy2012efficient}
Andr{\'a}s Gyorgy, Tam{\'a}s Linder, and G{\'a}bor Lugosi.
\newblock Efficient tracking of large classes of experts.
\newblock \emph{IEEE Transactions on Information Theory}, 58\penalty0
  (11):\penalty0 6709--6725, 2012.

\bibitem[Haussler et~al.(1998)Haussler, Kivinen, and Warmuth]{hauss98}
David Haussler, Jyrki Kivinen, and Manfred~K. Warmuth.
\newblock Sequential prediction of individual sequences under general loss
  functions.
\newblock \emph{IEEE Transactions on Information Theory}, 44\penalty0
  (5):\penalty0 1906--1925, 1998.

\bibitem[Hazan(2016)]{oco}
Elad Hazan.
\newblock Introduction to online convex optimization.
\newblock \emph{Foundations and Trends in Optimization}, 2\penalty0
  (3-4):\penalty0 157--325, 2016.

\bibitem[Hazan and Seshadhri(2009)]{hazan09}
Elad Hazan and Comandur Seshadhri.
\newblock Efficient learning algorithms for changing environments.
\newblock In \emph{Proceedings of the 26th annual international conference on
  machine learning}, ICML '09, pages 393--400, 2009.

\bibitem[Herbster and Warmuth(1998)]{hw98}
Mark Herbster and Manfred~K. Warmuth.
\newblock Tracking the best expert.
\newblock \emph{Machine Learning}, 32\penalty0 (2):\penalty0 151--178, August
  1998.

\bibitem[Jun et~al.(2017)Jun, Orabona, Wright, and Willett]{jun17a}
Kwang-Sung Jun, Francesco Orabona, Stephen Wright, and Rebecca Willett.
\newblock {Improved Strongly Adaptive Online Learning using Coin Betting}.
\newblock In \emph{Proceedings of the 20th International Conference on
  Artificial Intelligence and Statistics (AISTATS)}, volume~54, pages 943--951,
  2017.

\bibitem[Koolen and de~Rooij(2008)]{kdr08}
Wouter~M. Koolen and Steven de~Rooij.
\newblock Combining expert advice efficiently.
\newblock In \emph{Proceedings of the 21st Annual Conference on Learning Theory
  (COLT)}, pages 275--286, 2008.

\bibitem[Koolen and de~Rooij(2013)]{kdr13}
Wouter~M. Koolen and Steven de~Rooij.
\newblock Universal codes from switching strategies.
\newblock \emph{IEEE Transactions on Information Theory}, 59\penalty0
  (11):\penalty0 7168--7185, November 2013.

\bibitem[Koolen and van Erven(2015)]{squint}
Wouter~M. Koolen and Tim van Erven.
\newblock Second-order quantile methods for experts and combinatorial games.
\newblock In \emph{Proceedings of the 28th Annual Conference on Learning Theory
  (COLT)}, pages 1155--75, 2015.

\bibitem[Koolen et~al.(2012)Koolen, Adamskiy, and Warmuth]{kool}
Wouter~M. Koolen, Dmitry Adamskiy, and Manfred~K. Warmuth.
\newblock Putting bayes to sleep.
\newblock In \emph{Advances in Neural Information Processing Systems 25}, pages
  135--143. Curran Associates, Inc., 2012.

\bibitem[Koolen et~al.(2014)Koolen, van Erven, and Gr{\"u}nwald]{llr}
Wouter~M. Koolen, Tim van Erven, and Peter Gr{\"u}nwald.
\newblock Learning the learning rate for prediction with expert advice.
\newblock In \emph{Advances in Neural Information Processing Systems 27}, pages
  2294--2302. Curran Associates, Inc., 2014.

\bibitem[Luo and Schapire(2015)]{adanormalhedge}
Haipeng Luo and Robert~E. Schapire.
\newblock Achieving all with no parameters: Adaptive normalhedge.
\newblock In \emph{Proceedings of the 28th Annual Conference on Learning Theory
  (COLT)}, pages 1286--1304, 2015.

\bibitem[McQuade and Monteleoni(2012)]{mcquade2012global}
Scott McQuade and Claire Monteleoni.
\newblock Global climate model tracking using geospatial neighborhoods.
\newblock In \emph{AAAI}, 2012.

\bibitem[Merhav and Feder(1998)]{merhav}
Neri Merhav and Meir Feder.
\newblock Universal prediction.
\newblock \emph{IEEE Transactions on Information Theory}, 44:\penalty0
  2124--2147, 1998.

\bibitem[Monteleoni et~al.(2011)Monteleoni, Schmidt, Saroha, and
  Asplund]{monteleoni2011tracking}
Claire Monteleoni, Gavin~A Schmidt, Shailesh Saroha, and Eva Asplund.
\newblock Tracking climate models.
\newblock \emph{Statistical Analysis and Data Mining}, 4\penalty0 (4):\penalty0
  372--392, 2011.

\bibitem[Ryabko(1984)]{ryabko1984twice}
Boris~Y. Ryabko.
\newblock Twice-universal coding.
\newblock \emph{Problems of information transmission}, 20\penalty0
  (3):\penalty0 173--177, 1984.

\bibitem[Ryabko(1988)]{ryabko1988prediction}
Boris~Y. Ryabko.
\newblock Prediction of random sequences and universal coding.
\newblock \emph{Problems of information transmission}, 24\penalty0
  (2):\penalty0 87--96, 1988.

\bibitem[Shalev-Shwartz(2012)]{oloco}
Shai Shalev-Shwartz.
\newblock Online learning and online convex optimization.
\newblock \emph{Found. Trends Mach. Learn.}, 4\penalty0 (2):\penalty0 107--194,
  February 2012.

\bibitem[Shalizi et~al.(2011)Shalizi, Jacobs, Klinkner, and
  Clauset]{shalizi2011adapting}
Cosma~Rohilla Shalizi, Abigail~Z. Jacobs, Kristina~Lisa Klinkner, and Aaron
  Clauset.
\newblock Adapting to non-stationarity with growing expert ensembles.
\newblock \emph{arXiv preprint {\tt arXiv:1103.0949}}, 2011.

\bibitem[Shamir and Merhav(1999)]{shamir1999low}
Gil Shamir and Neri Merhav.
\newblock Low-complexity sequential lossless coding for piecewise-stationary
  memoryless sources.
\newblock \emph{IEEE transactions on information theory}, 45\penalty0
  (5):\penalty0 1498--1519, 1999.

\bibitem[Stoltz(2010)]{stoltz}
Gilles Stoltz.
\newblock Agr{\'e}gation s{\'e}quentielle de pr{\'e}dicteurs : m{\'e}thodologie
  g{\'e}n{\'e}rale et applications {\`a} la pr{\'e}vision de la qualit{\'e} de
  l'air et {\`a} celle de la consommation {\'e}lectrique.
\newblock \emph{Journal de la Soci{\'e}t{\'e} Fran{\c c}aise de Statistique},
  151\penalty0 (2):\penalty0 66--106, 2010.

\bibitem[Vovk(1998)]{vovk}
Vladimir Vovk.
\newblock A game of prediction with expert advice.
\newblock \emph{Journal of Computer and System Sciences}, 56\penalty0
  (2):\penalty0 153--173, 1998.

\bibitem[Vovk(1999)]{vovk99}
Vladimir Vovk.
\newblock Derandomizing stochastic prediction strategies.
\newblock \emph{Machine Learning}, 35\penalty0 (3):\penalty0 247--282, 1999.

\bibitem[Willems(1996)]{willems1996coding}
Frans M.~J. Willems.
\newblock Coding for a binary independent piecewise-identically-distributed
  source.
\newblock \emph{IEEE transactions on information theory}, 42\penalty0
  (6):\penalty0 2210--2217, 1996.

\bibitem[Wintenberger(2017)]{wintenberger2017boa}
Olivier Wintenberger.
\newblock Optimal learning with {Bernstein} online aggregation.
\newblock \emph{Machine Learning}, 106\penalty0 (1):\penalty0 119--141, 2017.

\end{thebibliography}

\appendix

\section{Proof of Proposition~\ref{prop:exp}}
%%% \section{Regret of the exponential weights algorithm}
\label{ap:prop1}

\begin{myproof}
  Since the loss function is $\eta$-exp-concave and $x_t = \sum_{i=1}^M v_{i,t}\, x_{i,t}$, we have
  \begin{equation*}
    e^{-\eta\, \ell(x_t,y_t)} \geq \sum_{i=1}^M v_{i,t}\, e^{-\eta\, \ell (x_{i,t}, y_t)} ,
    \qquad
    \mbox{\ie}
    \qquad
    \ell_{t} \leq - \frac 1\eta \log \left( \sum_{i=1}^M v_{i,t}\, e^{-\eta\, \ell_{i,t}} \right) .
  \end{equation*}
  This yields, introducing the posterior weights $v_{i,t}^m$ defined by~(\ref{eq:post}),
  \begin{equation*}
    \ell_t - \ell_{i,t}
    \leq - \frac 1\eta \log \left( \sum_{j=1}^M v_{j,t}\, e^{-\eta\, \ell_{j,t}} \right) - \ell_{i,t}
    = \frac 1\eta \log \left( \frac{e^{-\eta \, \ell_{i,t}}}{\sum_{j=1}^M v_{j,t}\, e^{-\eta \, \ell_{j,t}}} \right)
    = \frac 1\eta \log \frac{v_{i,t}^m}{v_{i,t}} \, .
  \end{equation*}
  Now recalling that the exponentially weighted average forecaster uses $\bm v_{t+1} = \bm v_t^m$, this writes: $\ell_t - \ell_{i,t} \leq \frac 1\eta \ln \frac{v_{i,t+1}}{v_{i,t}}$ which, summing over $t=1, \dots, T$, yields
  $L_T - L_{i,T} \leq \frac 1\eta \ln \frac{v_{i,T+1}}{v_{i,1}}$. Since $v_{i,1} = \pi_i$ and $v_{i,T+1} \leq 1$, this proves~(\ref{eq:reg-exp}); moreover, noting that $\ln \frac{v_{i,T+1}}{v_{i,1}} = \ln \frac{u_i}{v_{i,1}} - \ln \frac{u_i}{v_{i,T+1}}$, this implies
  \begin{equation*}
    \sum_{i=1}^M u_i\, ( L_T - L_{i,T})
    \leq \frac 1\eta \sum_{i=1}^M u_i\, \ln \frac{v_{i,T+1}}{v_{i,1}}
    = \frac 1\eta \big( \kll {\bm u}{\bm v_1} - \kll {\bm u}{\bm v_{T+1}} \big)
    \, ,
  \end{equation*}
  which establishes~(\ref{eq:reg-comb}) since $\bm v_1 = \bm \pi$ and $\kll {\bm u} {\bm v_{T+1}} \geq 0$.
\end{myproof}

\begin{rem}
  \label{rem:comb}
  We can recover the bound~(\ref{eq:reg-exp}) from inequality~(\ref{eq:reg-comb}) by considering $\bm u = \delta_i$. Conversely, inequality~(\ref{eq:reg-exp}) implies, by convex combination,
  \begin{equation*}
    L_T - \sum_{i=1}^M u_i\, L_{i,T}
    \leq  \frac 1\eta \sum_{i=1}^M u_i \log \frac 1{\pi_i} \, ;
  \end{equation*}
  inequality~(\ref{eq:reg-comb}) is actually an improvement on this bound, which replaces the terms $\ln \frac{1}{\pi_i}$ by $\ln \frac{u_i}{\pi_i}$.
  Following~\citet{kool}, this refinement is used in Section~\ref{sub:sleep} to obtain a tighter regret bound.
\end{rem}

\section{Proof of Theorem~\ref{thm:ghedge}}
\label{ap:thm-ghedge}

Theorem~\ref{thm:ghedge} is in fact a corollary of the more general Proposition~\ref{prop:gshedge}, valid in the specialist setting.

\begin{proposition}  
\label{prop:gshedge}
Assume we are given a set $\M$ of specialists, as well as a positive weight function $\pi : \M \to \R_+^*$. Assume that, at each time step $t\geq 1$, the set $A_t$ of active specialists is finite. Then, denoting $A_{\leq t} = \bigcup_{1\leq s \leq t} A_s$, the aggregation of specialists\footnote{Denoting, as in equation~(\ref{eq:shedge}), $L_{i,t} = \sum_{s\leq t \pp i \in A_s} \ell_{i,s} + \sum_{s \leq t \pp i \not\in A_s} \ell_s$ for each specialist $i$ and $t\geq 1$.}
\begin{equation}
  \label{eq:gshedge}
    x_t = \frac{\sum_{i \in A_t} \pi (i) \, e^{-\eta\, L_{i,t-1}} \, x_{i,t}}
    {\sum_{i \in A_t} \pi (i) \, e^{-\eta\, L_{i,t-1}}}
\end{equation}
    achieves the following regret bound: for each $T\geq 1$ and $i \in \M$, we have
\begin{equation}
  \label{eq:reg-gshedge}
      \sum_{t\leq T \pp i \in A_t} (\ell_{t}- \ell_{i,t})
      \leq \frac 1{\eta} \log
      \bigg( \frac{1}{\pi (i)} \sum_{j \in A_{\leq T}} \pi (j)  \bigg) \, .
\end{equation}
\end{proposition}

%\vspace{-2mm}
    \begin{myproof}
      {\bf\!{}of Proposition~\ref{prop:gshedge}\ \ }
      Fix $T \geq 1$, and denote $\Pi_T :=
%      \Pi (A_{\leq T}) :=
      \sum_{i \in A_{\leq T}} \pi (i)$. For $t = 1, \dots, T$, the forecast~(\ref{eq:gshedge}) may be rewritten as     
      \begin{equation*}
    x_t = \frac{\sum_{i \in A_t} \frac{\pi (i)}{\Pi_T} \, e^{-\eta\, L_{i,t-1}} \, x_{i,t}}
    {\sum_{i \in A_t} \frac{\pi (i)}{\Pi_T} \, e^{-\eta\, L_{i,t-1}}} 
  \end{equation*}
  which corresponds precisely to the aggregation of the set of specialists $A_{\leq T}$ with prior weights $\pi (i)/\Pi_T$ and active specialists $A_t \subset A_{\leq T}$ (up to time $T$). %Bound~
  (\ref{eq:reg-gshedge}) now follows from Proposition~\ref{prop:spec}.
    \end{myproof}

    \begin{myproof}
     {\bf\!{}of Theorem~\ref{thm:ghedge}\ \ }
     It suffices to notice that the weights of \ghedge\ are, for $i \leq M_t$, $w_{i,t} = \pi_i \, e^{-\eta L_{i,t-1}}$ with $L_{i,t-1} = L_{\tau_i-1} + \sum_{s=\tau_i}^t \ell_{i,s}$; hence, the forecasts of \ghedge\ are those of equation~(\ref{eq:gshedge}), and we can apply Proposition~\ref{prop:gshedge}.
   \end{myproof}

\section{Proof of Lemma~\ref{lem:seq1} and instantiations of algorithm \mhedge
%  and Corollaries~\ref{cor:fs} and~\ref{cor:ds}
}
\label{ap:prop2}

%% Here, we proceed to prove Lemma~\ref{lem:seq1}. In other words, we will show that the predictions of the exponential aggregation of sequences of experts with prior $\mu$, the Markov probability distribution with initial distribution $\bm \theta_1$ and transition matrices $\bm \theta_t$, are computed by algorithm \aggmarkov.

%%\bigskip

% also mention initialization ? trivial, but still

\begin{proof}%  
  {}{\bf of Lemma~\ref{lem:seq1}\ }
  Denote, for each $t \geq 1$, $\pi^t (i_1, \dots, i_t ) = \theta_1 (i_1) \, \theta_2 (i_2 \cond i_1) \cdots \theta_t (i_t \cond i_{t-1})$. Let $T \geq 1$ be arbitrary. We need to show that the predictions $x_t$ of the exponentially weighted aggregation of sequences of experts $i^T$ under the prior $\pi^T$ at times $t = 1, \dots, T$ coincide with those of algorithm \mhedge.

  \medskip
  First note that, by definition and since $L_{t-1} (i^T) = \sum_{s=1}^{t-1} \ell_{i_s,s} =: L_{t-1} (i^{t-1})$ does not depend on $i_t^T = (i_t, \dots, i_T)$, we have for $1\leq t \leq T$
  \begin{align*}
    x_t &= \frac{\sum_{i^T} \pi^T (i^T) \, e^{-\eta \, L_{t-1} (i^T)}\, x_t (i^T)}    
    {\sum_{i^T} \pi^T (i^T) \, e^{-\eta \, L_{t-1} (i^T)}}
    = \frac{\sum_{i^t,i_{t+1}^T} \pi^T (i^t,i_{t+1}^T) \, e^{-\eta \, L_{t-1} (i^{t-1})}\, x_{i_t,t}}    
    {\sum_{i^t,i_{t+1}^T} \pi^T (i^{t}, i_{t+1}^T) \, e^{-\eta \, L_{t-1} (i^{t-1})}} \\
    &\underset{(\star)}{=} \frac{\sum_{i^t} \pi^t (i^t) \, e^{-\eta \, L_{t-1} (i^{t-1})}\, x_{i_t,t}}    
      {\sum_{i^t} \pi^t (i^{t}) \, e^{-\eta \, L_{t-1} (i^{t-1})}}
      = \frac{\sum_{i^{t-1},i} \pi^t (i^{t-1},i) \, e^{-\eta \, L_{t-1} (i^{t-1})}\, x_{i,t}}    
      {\sum_{i^{t-1},i} \pi^t (i^{t-1},i) \, e^{-\eta \, L_{t-1} (i^{t-1})}}
  \end{align*}
  where $(\star)$ is a consequence of the identity $\sum_{i_{t+1}^T} \pi^T (i^t, i_{t+1}^T) = \pi^t (i^t)$.
  Hence, denoting  $w_t (i^t) := \pi^t (i^{t}) \, e^{-\eta L_{t-1}(i^{t-1})}$, we have
  \begin{equation*}    
    x_t
%    = \frac{ \sum_{i^t} w_t (i^t) \, x_t (i^t) }{ \sum_{i^t} w_t (i^t) }
    = \frac{\sum_{i} \sum_{i^{t-1}} w_t (i^{t-1},i) \, x_{i,t} }
     {\sum_{i} \sum_{i^{t-1}} w_t (i^{t-1},i) } 
    = \frac{\sum_{i=1}^M w_{i,t} \, x_{i,t}}{\sum_{i=1}^M w_{i,t}} 
    = \sum_{i=1}^M v_{i,t} \, x_{i,t}
\end{equation*}
where we set $w_{i,t} := \sum_{i^{t-1}} \pi^t (i^{t-1},i)\, e^{-\eta L_{t-1}(i^{t-1})}$ and $v_{i,t} := w_{i,t}/ \big(\!\sum_{j=1}^M w_{j,t} \big)$. To conclude the proof, it remains to show that the weights $\bm v_t$ are those computed by algorithm \mhedge.

\medskip
We proceed by induction on $t \geq 1$. For $t=1$, we have for every $i=1, \dots, M$, $w_{i,1} = w_1 (i) = \pi^1 (i) = \theta_1 (i)$ and hence $v_{i,1} = \theta_1 (i)$, \ie $\bm v_1 = \bm \theta_1$.
Moreover, for every $t\geq 1$, the identity $\pi^{t+1} (i^{t+1}) = \pi^t (i^t) \, \theta_{t+1} (i_{t+1} \cond i_t)$ implies
\begin{align*}
  w_{t+1} (i^{t+1}) &= \pi^{t+1} (i^{t+1}) \, e^{-\eta L_t (i^t)} \\
  &= \theta_{t+1} (i_{t+1} \cond i_t ) \, \pi^t (i^{t}) \, e^{-\eta L_{t-1} (i^{t-1})} \, e^{-\eta \, \ell_{i_t,t}} \\
  &= \theta_{t+1} (i_{t+1} \cond i_t) \, w_{t} (i^t) \, e^{-\eta \, \ell_{i_t,t}}
\end{align*}
\emph{i.e.}, for every $i,j$ and $i^{t-1}$,
$w_{t+1} (i^{t-1},j,i) = \theta_{t+1} (i \cond j) \, w_t (i^{t-1},j) \, e^{-\eta\, \ell_{j,t}}$. Summing over $i^{t-1}$ and $j$, this yields:
\begin{equation}
  \label{eq:proof-seqm1}
  w_{i,t+1} = \sum_{j=1}^M \theta_{t+1} (i \cond j) \, w_{j,t} \, e^{-\eta\, \ell_{j,t}} \, .
\end{equation}
Summing% equation
~(\ref{eq:proof-seqm1}) over $i=1, \dots, M$ gives $\sum_{i=1}^M w_{i,t+1} = \sum_{j=1}^M w_{j,t} \, e^{-\eta \, \ell_{j,t}}$ 
(since $\sum_{i=1}^M \theta_{t+1} (i \cond j) =~1$)
and therefore
\begin{equation*}
  v_{i,t+1}
  = \frac{w_{i,t+1}}{\sum_{j=1}^M w_{j,t+1}}
  = \frac{\sum_{j=1}^M \theta_{t+1} (i \cond j) \, w_{j,t} \, e^{-\eta\, \ell_{j,t}}}
  {\sum_{j=1}^M w_{j,t} \, e^{-\eta \, \ell_{j,t}}} 
  = \frac{\sum_{j=1}^M \theta_{t+1} (i \cond j) \, v_{j,t} \, e^{-\eta\, \ell_{j,t}}}
  {\sum_{j=1}^M v_{j,t} \, e^{-\eta \, \ell_{j,t}}}
  = \sum_{j=1}^M \theta_{t+1} (i \cond j) \, v_{j,t}^m 
\end{equation*}
where $\bm v_{t}^m$ is the posterior distribution, defined by equation~(\ref{eq:post}). 
This corresponds precisely to the update of the \aggmarkov{} algorithm, which completes the proof.
\end{proof}

%%\paragraph{Notable examples.}
 We now instantiate the generic algorithm~\aggmarkov{} 
and Proposition~\ref{prop:mhedge} on specific choices of  prior weights  and transition probabilities. This enables to recover a number of results from the literature. For concreteness, we take $\bm \theta_1 = \frac 1M \bm 1$.

\begin{corollary}[Fixed share]
  \label{cor:fs}
Setting $\theta_t (i\cond j) = (1-\alpha)\, \bm 1_{i=j} + \alpha\, \frac 1M$ with $\alpha \in (0,1)$, this leads to the Fixed-Share algorithm of \citet{hw98} with update
$\bm v_{t+1} = (1- \alpha)\, \bm v_{t}^m + \alpha \, \frac 1M \bm 1$ 
and regret bound
\begin{equation}
  \label{eq:reg-fs1}
    \sum_{t=1}^T \ell_t - \sum_{t=1}^T \ell_{i_t,t}
    \leq
    \frac{k+1}\eta \log M
    + \frac{k}{\eta} \ln \frac 1\alpha
    + \frac{T-k-1}{\eta} \ln \frac 1{1-\alpha}
    \, ,
  \end{equation}  
  where $k = k(i^T)$ denotes the number of shifts, $1 < \sigma_1 < \dots < \sigma_k \leq T$ these shifts (such that $i_{\sigma_j} \neq i_{\sigma_j -1}$) and $\sigma_0 =1$. When $T$ and $k$ are fixed and known, this bound is minimized by choosing $\alpha = \frac{k}{T-1}$ and becomes, denoting $H (p) = -p \ln p - (1-p) \ln (1-p)$ the binary entropy function,
  \begin{equation}
    \label{eq:reg-fs2}
    \frac{k+1}{\eta} \ln M + \frac{T-1}\eta H \Big( \frac k{T-1} \Big)
%%%  \frac{k+1}{\eta} \ln M + \frac k\eta \ln \frac{T-1}k + \frac{T-k-1}\eta \ln \frac{T-1}{T-k-1}
    \leq \frac{k+1}{\eta} \ln M + \frac k\eta \ln \frac{T-1}k + \frac k\eta  \, .
  \end{equation}
\end{corollary}

\begin{rem}
The quantity of equation~(\ref{eq:reg-fs2}), \ie the bound on the regret of fully tuned Fixed Share algorithm, is essentially equal to the optimal bound $\frac 1\eta \log \binom{T-1}{k} M^{k+1} \approx \frac{k+1}{\eta} \log M + \frac{k}{\eta} \log \frac{T-1}{k}$, obtained by aggregating all sequences of experts with at most $k$ shifts (which would require to maintain a prohibitively large number of weights).
\end{rem}

% The main shortcoming of the Fixed Share algorithm is that the parameter $\alpha$ needs to be set in advance, which is problematic when the comparator parameters $T$ and $k$ are unknown. It turns out that one can do almost as well, by using time-varying transition probabilities.

\begin{corollary}[Decreasing share]
  \label{cor:ds}
%  Let $(\alpha_t)_{t\geq 2}$ be a decreasing sequence in $[0,1]$. 
  Consider the special case of algorithm~\aggmarkov\ where $\theta_t (i\cond j) = (1-\alpha_t)\, \bm 1_{i=j} +  \frac {\alpha_t}M $, so that the update becomes
  $\bm v_{t+1} = (1-\alpha_{t+1})\, \bm v_{t}^m + \frac{\alpha_{t+1}}{M} \, \bm 1$. For every $T \geq 1$, $0 \leq k \leq T$, and every sequence of experts $i^T = (i_1 , \dots, i_T)$ with $k$ shifts at times $\sigma_1 < \dots < \sigma_k$,
  % it achieves the following:
  \begin{equation}
    \label{eq:reg-ds1}
      \sum_{t=1}^T \ell_t - \sum_{t=1}^T \ell_{i_t,t}
      \leq
      \frac{k+1}\eta \ln M + \frac 1\eta \sum_{j=1}^k \ln \frac 1{\alpha_{\sigma_j}}
      + \frac 1\eta \sum_{t = 2}^T \ln \frac{1}{1-\alpha_{t}}
  \end{equation}
  In the special case\footnote{Which we consider because of the simplicity of the bound as well as its proof, involving a %nice
    telescoping simplification; it is akin to Theorem~10 of~\citet{kdr13}, which uses $\alpha_t = 1- e^{-c/t}$.} when $\alpha_t = \frac 1t$, this bound becomes, for \emph{every} $T$, $k$ and $i^T$:
     \begin{equation}
      \label{eq:reg-ds2}
      \sum_{t=1}^T \ell_t - \sum_{t=1}^T \ell_{i_t,t}
      \leq \frac{k+1}\eta \ln M + \frac 1\eta \sum_{j=1}^k \ln \sigma_j + \frac 1\eta \log T
      \leq \frac{k+1}\eta \ln M + \frac{k+1}\eta \ln T
      \, .
    \end{equation}
\end{corollary}

\begin{rem}
The result of Corollary~\ref{cor:ds} is worth emphasizing: at no computational overhead, the use of decreasing transition probabilities gives a bound essentially in $\frac 1\eta (k+1) \log M + \frac 1\eta k \log T$ valid for every $T$ and $k$, which is close to the bound $\frac 1\eta (k+1) \log M + \frac 1\eta k \log \frac Tk$ one gets by optimally tuning $\alpha$ as a function of $T$ and $k$ in the Fixed Share algorithm, particularly when $k \ll T$ (in this latter case of rare shifts, the first, sharper bound of equation~(\ref{eq:reg-ds2}) is even more appealing).
\end{rem}

\begin{myproof}\textbf{of corollaries~\ref{cor:fs} and~\ref{cor:ds}\ }
  We consider the Decreasing Share algorithm, with time-varying transition probabilities $\alpha_t \in (0,1)$ (the Fixed Share algorithm corresponds to the special case $\alpha_t = \alpha$).
  Let $i^T = (i_1, \dots, i_T)$ be a sequence of experts with shifts at times $\sigma_1 < \dots < \sigma_k$. 
  By Proposition~\ref{prop:mhedge}, we have
    \begin{align*}
    \sum_{t=1}^T \ell_t - \sum_{t=1}^T \ell_{i_t,t}
      &\leq \frac 1\eta \log \frac 1{1/M}
        + \frac 1\eta \sum_{j=1}^k \ln \frac 1{\alpha_{\sigma_j}/M}
      + \frac 1\eta \sum_{t \neq \sigma_j} \ln \frac{1}{1-\alpha_{\sigma_j} + \alpha_{\sigma_j}/M} \\
      &\leq \frac{k+1}\eta \ln M
        + \frac 1\eta \sum_{j=1}^k \ln \frac 1{\alpha_{\sigma_j}}
        + \frac 1\eta \sum_{t \neq \sigma_j} \ln \frac{1}{1-\alpha_{\sigma_j}}
  \end{align*}
  Corollary~\ref{cor:fs} directly follows by taking $\alpha_t = \alpha$ in the above inequality, whereas the bound~(\ref{eq:reg-ds1}) of Corollary~\ref{cor:ds} is obtained by bounding
  $\sum_{t \neq \sigma_j} \log \frac 1{1-\alpha_t} \leq \sum_{t=2}^T \log \frac 1{1-\alpha_t}$. In the case when $\alpha_t = \frac 1t$, we recover~(\ref{eq:reg-ds2}) by substituting for $\alpha_t$ and noting that
  \begin{equation}
    \label{eq:tel}
    \sum_{t=2}^T \log \frac 1{1-1/t} = \sum_{t=2}^T \log \frac{t}{t-1}
    = \log T \, .
  \end{equation}
\end{myproof}

\vspace{-2mm}
\section{Proof of Proposition~\ref{prop:reg-sleep}}
\label{ap:sleep}

\begin{myproof}
  Since $x_t (i,1) = x_{i,t}$ and $x_{t} (i,0) = x_t$,
  equation~(\ref{eq:pred-sleep}) implies that the forecast $x_t$ of \smhedge{} satisfies:
  \vspace{-1mm}
  \[ x_t = \sum_{i=1}^M \sum_{a\in \{ 0,1 \}} v_t (i,a) \, x_t (i,a) \, . \] Hence, \smhedge\ reduces to algorithm \mhedge\ over the sleeping experts, \ie  (by Lemma~\ref{lem:seq1}, up to time $T$) to the exponentially weighted aggregation of sequences of sleeping experts under the Markov prior $\pi ((i,a_t)_{1\leq t \leq T}) = \theta_{i,1} (a_1) \prod_{t=2}^T \theta_{i,t} (a_{t} \cond a_{t-1})$ (and $0$ for other sequences)% up to time $T$
  . Hence, if $u$ is the uniform probability on the $n$ sequences $(e_p , a_{p,t})_{1\leq t \leq T}$, $1\leq p \leq n$, we have by Proposition~\ref{prop:exp}:
  \vspace{-1mm}
  \begin{equation}
    \label{eq:sleep-general1}
    \sum_{\iota^T} u (\iota) \, (L_T - L_T (\iota^T))
    \leq \frac 1\eta \kll u \pi
    = \frac 1{\eta} \frac 1n
    \sum_{p=1}^n \log \frac{1/n}{\pi ((e_p , a_{p,t})_{1\leq t \leq T})}
  \end{equation}
%  \vspace{-1mm}
%%%  the bound~(\ref{eq:reg-sleep}) follows by substituting for $\pi$ in the right-hand side, and by recalling that the left-hand side equals $\frac 1n (L_T - L_T (i^T))$ by the previous decomposition.
  As shown in the reformulation of the regret with respect to sparse sequences of experts of Section~\ref{sub:sleep}, the left hand side of equation~\eqref{eq:sleep-general1} equals $\frac 1n (L_T - L_T (i^T))$. The desired regret bound~\eqref{eq:reg-sleep} follows by substituting for $\pi$ in the right-hand side.
\end{myproof}

\section{Uniform bounds and optimality}
% Comparison with information-theoretic bounds
\label{ap:info}

In this section, we provide simple bounds derived from Theorems~\ref{thm:ghedge}, \ref{thm:fmhedge}, \ref{thm:gmhedge} and \ref{thm:gsmhedge} that are not quite as adaptive to the parameters of the comparison class as the ones provided in Section~\ref{sec:overview}, but are more uniform and hence more interpretable.
We then discuss the optimality of these bounds, by relating them either to theoretical lower bounds or to information-theoretic upper bounds (obtained by naively aggregating all elements of the comparison class, which is computationally prohibitive).
% We then relate those bounds to the information-theoretic bounds, obtained by naively aggregating all elements of the comparison class.

\paragraph{Constant experts}

Consider the algorithm~\ghedge{}
% and \fmhedge{},
with the uniform (unnormalized) prior: $\pi_i = 1$ for each $i \geq 1$. By Theorem~\ref{thm:ghedge}, %and~\ref{thm:fmhedge},
this algorithm achieves the regret bound
\[ \frac 1 \eta \log M_T \]
with respect to each constant expert.

\smallskip
This regret bound cannot be improved in general: 
% , even knowing $T$ and $M_T$:
indeed, consider the logarithmic loss on $\N^*$, defined by $\ell (x, y) = - \log x (y)$ for every $y \in \N^*$ and every probability distribution $x$ on $\N^*$. Fix $T \geq 1$, and consider the sequence $y_t = x_{i,t} = 1$ ($1 \leq t < T$, $1\leq i \leq M_t$) and $y_t \in \{ 1, \dots, M_T \}$ and $x_{i,T} = i$ for $i = 1, \dots, M_T$.
For each $i = 1, \dots, M_T$, we have $\sup_{1\leq i \leq M_T} (L_T - L_{i,T}) = \sup_{1\leq i \leq M_T}  - \log \frac{x_t(y_t)}{x_{i,t} (y_t)}  = - \log x_t(y_t)$. Now whatever $x_t$ is, there exists $y_t \in \{ 1, \dots, M_T \}$ such that $x_t (y_t) \leq \frac 1{M_T}$ (since $x_t$ sums to $1$). Since $y_t$ is picked by an adversary after $x_t$ is chosen, the adversary can always ensure a regret of at least $\log M_T$.

\paragraph{Fresh sequences of experts}
By Theorems~\ref{thm:ghedge} and~\ref{thm:fmhedge}, algorithms \ghedge{} and \fmhedge{} with a uniform prior ($\pi_i = 1$ for each $i \geq 1$) achieve the regret bound
\begin{equation*}
  L_T -
  L_T (i^T)  
  \leq  
  \frac 1\eta \sum_{j=1}^k \log M_{\sigma_{j}-1}  
  + \frac 1\eta \log {M_T}  
\end{equation*}
for every sequence $i^T$ of fresh experts with shifts at times $\bm \sigma = (\sigma_1, \dots, \sigma_k)$. By the same argument as above, this bound cannot be improved in general.

\paragraph{Arbitrary admissible sequences of experts}

By Theorem~\ref{thm:gmhedge}, algorithm~\gmhedge{} with uniform prior $\bm \pi$ and transition probabilities $\alpha_t = \frac 1t$ achieves, for every admissible sequence $i^T$
\[
      L_T - L_T (i^T)
  \leq
  \frac 1\eta 
    \sum_{j=0}^{k} \log M_{\sigma_{j+1}-1}
    +   \frac 1\eta \sum_{j=1}^{k_1} \log {\sigma^1_j}
    +   \frac 1\eta \log T
    \leq \frac 1\eta (k+1) \log M_T + \frac 1\eta (k_1 +1)\log T
    \, .
    \]
  where  $\bm \sigma^0 = (\sigma^0_1 , \dots, \sigma^0_{k_0})$ (resp. $\bm \sigma^1 = (\sigma^1_1 , \dots, \sigma^1_{k_1})$) denotes the shifts to fresh  (resp. incumbent) experts, with $k=k_0+k_1$.

  \smallskip
  This simple bound is close to the information-theoretic bound obtained by aggregating all admissible sequences of experts: indeed, the number of such sequences is bounded by (with equality if $M_T = M_1$) $M_T^{k+1} \binom{T-1}{k_1}$ (an admissible sequence is determined by its switches to fresh experts -- at most $M_T^{k_0 +1}$ possibilities -- and its switches to incumbent experts -- at most $M_{T}^{k_1}$ possibilities for the choices of the experts, and at most $\binom{T-1}{k_1}$ choices for the switches to incumbent experts).
  The regret bound corresponding to the aggregation of this large expert class is therefore of order
  \[ \frac 1\eta \log M_T^{k+1} \binom{T-1}{k_1} \approx \frac 1\eta (k+1) \log M_T + \frac 1\eta k_1 \log \frac{T-1}{k_1} \, , \]
  which is close to the bound of \ghedge, especially if $k_1 \ll T$.

  \paragraph{Sparse admissible sequences}

Finally, Theorem~\ref{thm:gsmhedge} implies that algorithm \gsmhedge, with uniform weights $\bm \pi$ and transition probabilities $\alpha_t = \beta_t = \frac 1{t \ln t}$, has a regret bound of 
\[
      L_T  - L_T (i^T)
%       \leq
%      \frac 1\eta n \ln \frac{M_T}{n}
%        + \frac 1\eta n \log (2T)  
%        + \frac 2\eta \sum_{j=1}^k \log \sigma_j
        \leq
              \frac 1\eta n \ln \frac{M_T}{n}
        + \frac 1\eta  n ( \log 2 +  c_T \ln \ln T)
        + \frac 2\eta k \log T + \frac 1\eta 2 k \log \log T \, .
      \]
      for any sparse admissible sequence $i^T$ with at most $k$ shifts and taking values in a pool of $n$ experts, where $c_T := (\ln \ln T)^{-1} \sum_{t = 2}^T \ln \frac{1}{1-\alpha_t} \longrightarrow_{T\to \infty} 1$.
      Again, for $k \ll T$, this is close to the information-theoretic upper bound obtained by aggregating all sparse sequences with $k$ shifts in a pool of $n$ experts, of approximately $ n \log \frac{M_T}n + (k+1) \log n + k \log \frac{T}{k}$. The main difference, namely the doubling of the term $k \log T$ in the regret bound of \gsmhedge, is not specific to the growing experts setting, and also appears in the context of a fixed set of experts~\citep{mpp,kool}.
      %; it has been interpreted by~\cite{kool} as a \emph{price of coordination}

\end{document}